%% file: main.tex
\documentclass[10pt,twocolumn,letterpaper]{article}

\usepackage{cvpr}
\usepackage{url}
\usepackage{booktabs}
\usepackage{multirow}
\usepackage{makecell}
\usepackage{graphicx}
\usepackage{caption}
\usepackage{amsmath}
\usepackage{amssymb}
\usepackage{pifont}
\usepackage{listings}
\usepackage{amsthm}
\usepackage{enumitem}
\usepackage{colortbl}
\usepackage{algorithmicx, algorithm}
\usepackage{algpseudocode}
\usepackage[accsupp]{axessibility} 
\usepackage[hypcap=false]{caption}

\input{preamble}

\input{notations.tex}

\algrenewcommand\algorithmicrequire{\textbf{Input:}}
\algrenewcommand\algorithmicensure{\textbf{Output:}}

\newcommand\blfootnote[1]{%
  \begingroup\renewcommand\thefootnote{}\footnote{#1}%
  \addtocounter{footnote}{-1}\endgroup}

\definecolor{cvprblue}{rgb}{0.21,0.49,0.74}
\usepackage[hypertexnames=false,pagebackref,breaklinks,colorlinks,allcolors=cvprblue]{hyperref}
\makeatletter
\AtBeginDocument{\renewcommand*{\theHALG@line}{\thealgorithm.\arabic{ALG@line}}}
\makeatother

\def\paperID{30848}
\def\confName{CVPR}
\def\confYear{2025}

\title{\algoname{}: Dynamic Patchification for Efficient Autoregressive Visual Generation}

\author{
    Divyansh Srivastava\textsuperscript{1$\dagger$} \quad
    Akshay Mehra\textsuperscript{2$*$} \quad
    Pranav Maneriker\textsuperscript{2$*$} \quad
    Debopam Sanyal\textsuperscript{3$\dagger$} \\
    Vishnu Raj\textsuperscript{2} \quad
    Vijay Kamarshi\textsuperscript{2} \quad
    Fan Du\textsuperscript{2} \quad
    Joshua Kimball\textsuperscript{2} \\
    \textsuperscript{1}University of California San Diego \quad
    \textsuperscript{2}Dolby Laboratories \quad
    \textsuperscript{3}Georgia Tech \quad
    \textsuperscript{$*$}Equal Contribution
}

\begin{document}

\twocolumn[{%
            \renewcommand\twocolumn[1][]{#1}%
            \maketitle

            \begin{center}
                \begin{minipage}{0.64\linewidth}
                    \centering
                    \includegraphics[width=\linewidth,trim=1mm 2mm 1mm 0mm,clip]{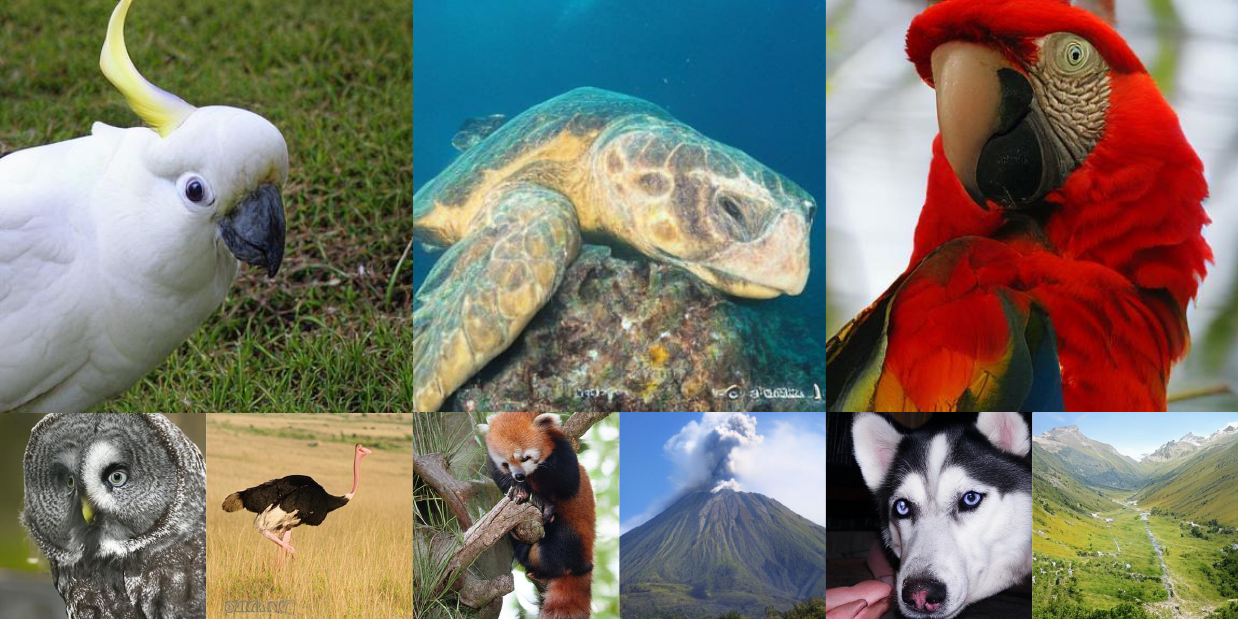}
                    {\small (a)}
                    \label{fig:teaser_samples}
                \end{minipage}
                \hfill
                \begin{minipage}{0.35\linewidth}
                    \centering
                    \includegraphics[width=\linewidth,trim=1mm 2mm 1mm 0mm,clip]{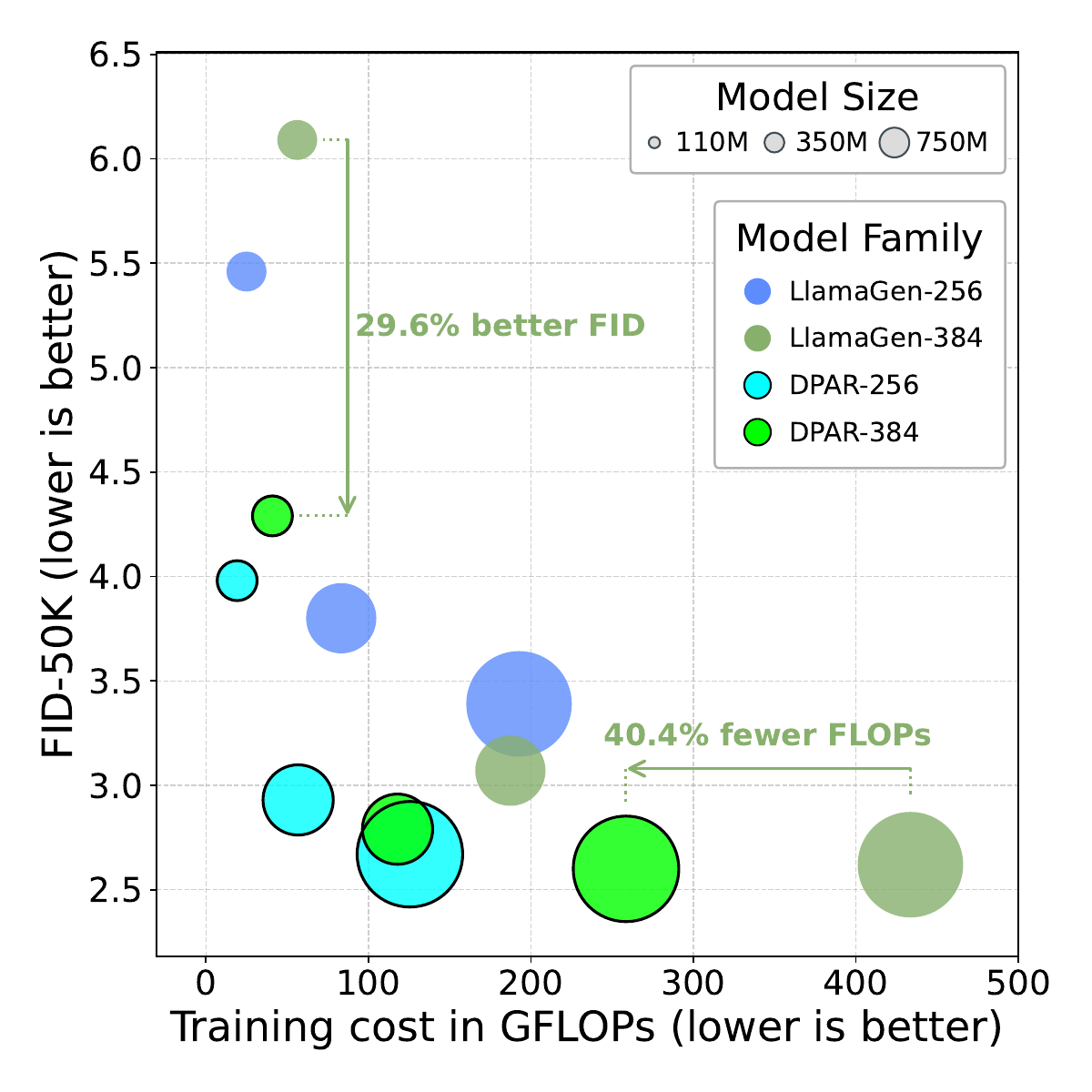}
                    {\small (b)}
                \end{minipage}

                \captionof{figure}{
                    \textbf{Autoregressive image generation with \algoname{}.}
                    (a) Selected samples from our class-conditional \algoname{}-384-XL model trained on ImageNet.
                    (b) FID vs.\ FLOPs for \algoname{} variants at $256 \times 256$ and $384 \times 384$ resolution.
                    \algoname{} reduces training FLOPs by up to $40.4\%$ while improving FID by up to $29.6\%$ over baseline models.
                }
                \label{fig:teaser}
            \end{center}
        }]

\blfootnote{$\dagger$ Work done during internship at Dolby Laboratories.}
\blfootnote{Code: \url{https://github.com/dsrivastavv/DPAR}}

\setlength{\parskip}{-1pt}
\input{sec/0_abstract}
\input{sec/1_introduction}
\input{images/methodology.tex}
\input{sec/2_related_work}

\input{sec/3_methodology}

\input{sec/4_experiments}

\input{sec/5_conclusion}
\clearpage
{
    \small
    \bibliographystyle{ieeenat_fullname}
    \bibliography{main}
}
\input{sec/X_supplementary}

\end{document}

%% file: preamble.tex
\usepackage{float}

%% file: notations.tex
\newcommand{\algoname}{DPAR}
\newcommand{\Itok}{I_{\mathrm{tok}}}
\newcommand{\Ipatch}{I_{\mathrm{patch}}}
\newcommand{\Hth}{\mathcal{H}_{\mathrm{Th}}}
\newcommand{\Hthtrain}{\mathcal{H}_{\mathrm{Th}}^{\mathrm{train}}}
\newcommand{\Hthinf}{\mathcal{H}_{\mathrm{Th}}^{\mathrm{inf}}}
\newcommand{\Pmax}{P_{\mathrm{max}}}
\newcommand{\Pavg}{P_{\mathrm{avg}}}
\newcommand{\Pstatic}{P_{\mathrm{static}}}
\newcommand{\Etokseq}{\mathbf{E}^{\mathrm{tok}}}
\newcommand{\Htokseq}{\mathbf{H}^{\mathrm{tok}}}
\newcommand{\Hpatchseq}{\mathbf{H}^{\mathrm{patch}}}
\newcommand{\Hpatchpred}{\widehat{\mathbf{H}}^{\mathrm{patch}}}
\newcommand{\Ztokseq}{\mathbf{Z}^{\mathrm{tok}}}
\newcommand{\Zpatchseq}{\mathbf{Z}^{\mathrm{patch}}}
\newcommand{\Htokmid}{\widetilde{\mathbf{H}}^{\mathrm{tok}}}
\newcommand{\Htokout}{\bar{\mathbf{H}}^{\mathrm{tok}}}
\newcommand{\Ptokpred}{\widehat{\mathbf{P}}^{\mathrm{tok}}}
\newcommand{\SA}{\operatorname{SA}}
\newcommand{\CA}{\operatorname{CA}}
\newcommand{\NormOp}{\operatorname{Norm}}
\newcommand{\LinearOp}{\operatorname{Linear}}

%% file: sec/0_abstract.tex
\begin{abstract}
Decoder-only autoregressive image generation typically relies on fixed-length tokenization schemes whose token counts grow quadratically with resolution, substantially increasing the computational and memory demands of attention. We present \textbf{\algoname{}}, a novel decoder-only autoregressive model that dynamically aggregates image tokens into a variable number of patches for efficient image generation. Our work is the first to demonstrate that next-token prediction entropy from a lightweight and unsupervised autoregressive model provides a reliable criterion for merging tokens into larger patches based on information content. \algoname{} makes minimal modifications to the standard decoder architecture, ensuring compatibility with multimodal generation frameworks and allocating more compute to the generation of high-information image regions. Further, we demonstrate that training with dynamically sized patches yields representations that are robust to patch boundaries, allowing \algoname{} to scale to larger patch sizes at inference. \algoname{} reduces token count by $1.81\times$ and $2.06\times$ on ImageNet 256 and 384 generation resolution, respectively, leading to a reduction of up to 40.4\% FLOPs in training costs. Further, our method exhibits faster convergence and improves FID by up to 29.6\% relative to baseline models.
\end{abstract}

%% file: sec/1_introduction.tex
\section{Introduction}

The large-scale success of autoregressive (AR) decoder-only language models~\citep{transformer,gpt1,brown2020language,openai2023gpt,bert,t5,llama1,touvron2023llama,ouyang2022training,kaplan2020scaling,wei2022chain,bai2023qwen,wu2025qwen,liu2024deepseek} has sparked a growing interest in extending next-token prediction paradigm to image generation, for a seamless integration with language models for unified multimodal generation~\citep{deng2025emergingpropertiesunifiedmultimodal,chameleonteam2025chameleonmixedmodalearlyfusionfoundation,wang2024emu3,xie2024show,zhou2024transfusion,yu2023scalingautoregressivemultimodalmodels,ge2023plantingseedvisionlarge}. Recent works~\citep{li2024autoregressive,sun2024autoregressive,pang2025randar,yu2024randomized,ramesh2021zero,yu2022scaling,tang2024hart} demonstrate that AR approaches can achieve performance comparable to, and in some cases surpass, diffusion models~\citep{scorebased,ddpm,ddim,adm,cfg,ldm,dalle2,imagen,esser2024scalingrectifiedflowtransformers,Srivastava_2025_ICCV}, which have long dominated the image generation~\citep{gan,pix2pix,stylegan,stylegan2,stylegan3,stylegan-xl,stylegan-t,gigagan,cyclegan,biggan} landscape. Typical AR image generation methods, including LlamaGen~\citep{sun2024autoregressive}, employ VQ-VAE~\citep{oord2018neuraldiscreterepresentationlearning,razavi2019generating,sun2024autoregressive} to tokenize images into discrete 2D tokens that are flattened into 1D sequences, followed by next-token prediction training with minimal changes to a decoder-only transformer model. However, a fundamental scalability challenge persists: the number of tokens scales quadratically with image resolution, leading to substantial increases in computational and memory costs for attention~\citep{tay2022efficient,dao2022flashattention,zhai2022scaling}. For instance, generating a 256×256 image with a standard $16\times$ downsampling requires generating 256 tokens, whereas a 1024×1024 image requires 4096 tokens — a $16\times$ increase in token count and context length for attention.

Existing methods have attempted to reduce token counts for AR image generation through 1D tokenization~\citep{shen2025cat,yu2024imageworth32tokens,duggal2024adaptive,miwa2025onedpieceimagetokenizermeets} and token compression techniques~\citep{havtorn2023msvit,ma2025token,bolya2023token,rao2021dynamicvitefficientvisiontransformers,liang2022not,fayyaz2022adaptive,kong2022spvit,ryoo2106tokenlearner,beyer2023flexivit,dehghani2023patch}. While 1D tokenizers reduce the number of tokens, they are often not favored due to the loss of 2D spatial structure, which is essential for zero-shot editing capabilities such as extrapolation and outpainting~\citep{pang2025randar}. Moreover, token compression methods typically merge statically by a fixed factor, often combining high-information regions, leading to information loss and degraded generation quality. In this work, we ask the question: \textit{Can we dynamically merge image tokens based on their information content while preserving the 2D spatial structure for efficient AR image generation?}

To this end, we propose \textbf{\algoname{}}, a novel autoregressive image generation model that dynamically aggregates discrete 2D image tokens into a variable number of \emph{patches} for efficient visual generation. Images often contain low-information regions, such as homogeneous areas like sky or walls, that can be represented with fewer tokens without information loss~\citep{he2022masked}. Inspired by recent advances in NLP~\citep{pagnoni2024bytelatenttransformerpatches,nawrot2022hierarchicaltransformersefficientlanguage,Nawrot_2023,sukhbaatar2019adaptive}, we leverage next-token prediction entropy from a lightweight unsupervised AR model to merge tokens into larger units called \emph{patches} (see~\Cref{fig:entropy_demo}). Image regions with homogeneous details have few potential choices and patterns for the next token, and vice versa, leading to low and high next-token prediction entropy, respectively. This allows merging tokens in low-information regions while preserving token-level granularity in high-information areas, resulting in a more balanced allocation of compute during generation. Overall, our method unifies the strengths of 2D tokenization, 1D tokenization, and token-merging approaches by preserving 2D spatial structure while dynamically merging tokens based on their information content.

Our method leverages a lightweight encoder that aggregates tokens into patches based on next-token prediction entropy, and a corresponding decoder that reconstructs tokens from generated patches. The autoregressive transformer operates on a reduced number of patches instead of tokens, reducing the computational cost of attention. We evaluate our proposed method on the ImageNet-256~\citep{russakovsky2015imagenetlargescalevisual} class-conditional image generation benchmark and demonstrate that \algoname{} achieves a $1.81\times$ and $2.06\times$ reduction in token count for 256×256 and 384×384 image generation, respectively, leading to a significant reduction of up to 40.4\% GFLOPs in training costs and improves FID by up to 29.6\% relative to baselines. Finally, we show that \algoname{}'s training with a dynamic patch-based representation yields robust representations, enabling \algoname{} to scale to larger patch sizes at inference. Our contributions are summarized below:
\begin{itemize}
    \item We present \textbf{\algoname{}}, a novel autoregressive image generation model that dynamically aggregates image tokens into a variable number of patches based on their information content, enabling efficient generation.
    \item \algoname{} achieves $1.81\times$ and $2.06\times$ reduction in token count on ImageNet 256×256 and 384×384 generation, respectively, leading to a significant reduction of up to 40.4\% GFLOPs in training costs. Further, our method exhibits faster convergence and improves FID by up to 29.6\% relative to baseline models.
    \item We demonstrate that \algoname{}'s training with dynamically sized patches makes its learned representations robust to patch boundaries, enabling \algoname{} to scale to larger patch sizes during inference for further efficiency gains. 
\end{itemize}

\input{images/entropy_study.tex}

%% file: images/entropy_study.tex
\begin{figure}[t]
    \centering
    \includegraphics[width=0.9\columnwidth]{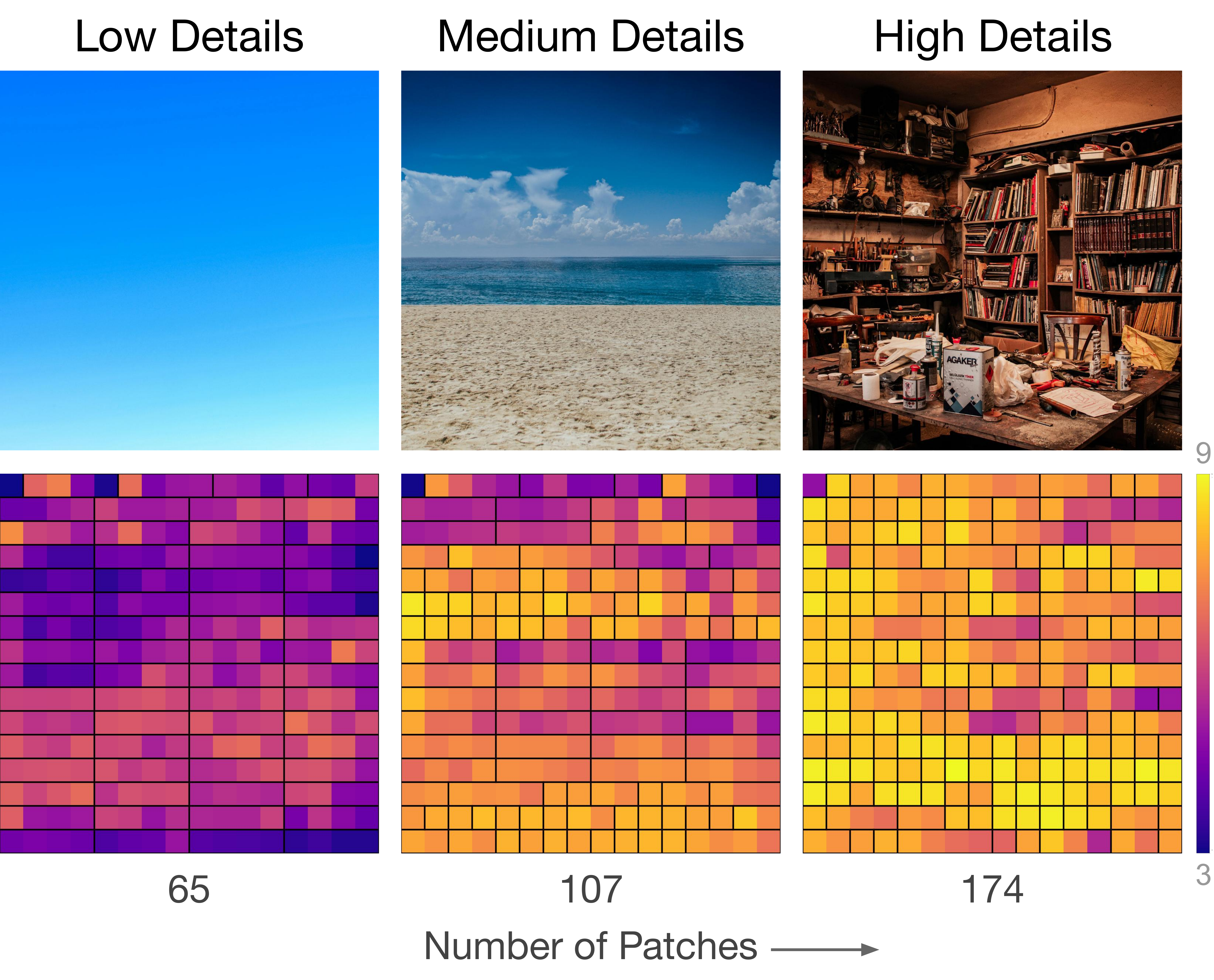}
    \caption{\textbf{Entropy-guided Token Aggregation.} \textit{(Top)} Input images with increasing information content. \textit{(Bottom)} Corresponding entropy heatmaps, where warmer colors indicate higher next-token prediction entropy. Low-information regions yield low-entropy tokens that are merged into patches, while high-information regions retain fine-grained token-level granularity. Black outlines denote final patch boundaries for 256$\times$256 images.}
    \label{fig:entropy_demo}
\end{figure}

%% file: images/methodology.tex
\begin{figure*}[t]
    \centering
    \includegraphics[width=0.9\textwidth]{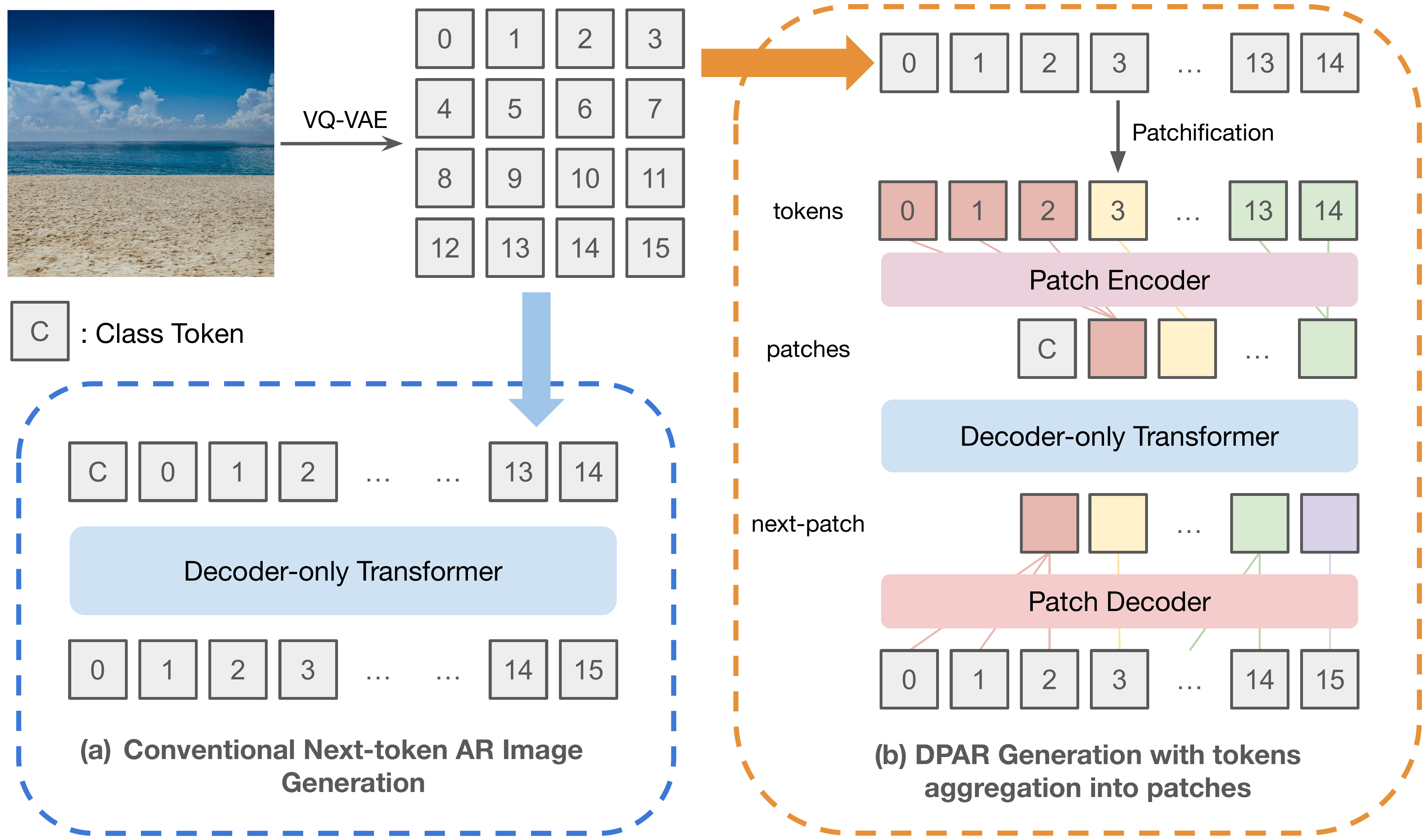}
    \caption{\textbf{Overview of \algoname{}}. (a) Conventional AR image generation employs decoder-only transformers that operate on a fixed number of tokens per image, where the token count scales quadratically with image resolution. (b) \algoname{} dynamically aggregates image tokens based on information content, producing a variable number of patches per image. Decoder-only transformers then operate on this reduced set of patches, lowering both computational and memory overhead of attention. \algoname{} requires minimal modifications to the standard decoder architecture, preserving compatibility with multimodal generation frameworks.}
    \label{fig:methodology}
\end{figure*}

%% file: sec/2_related_work.tex
\section{Related Work}

\paragraph{Autoregressive Image Generation} Seminal work LlamaGen~\citep{sun2024autoregressive} adopts a decoder-only Llama~\citep{llama1} architecture and is trained with standard next-token prediction (NTP) loss on discrete 2D VQ-VAE~\citep{oord2018neuraldiscreterepresentationlearning} tokens rasterized left-to-right to 1D sequence, achieving performance comparable to diffusion models~\citep{scorebased,ddpm,ddim,adm,cfg,ldm,dalle2,imagen,esser2024scalingrectifiedflowtransformers,Srivastava_2025_ICCV}. Later works have proposed improvements to vanilla AR image generation along two major directions: i) \textbf{Random-order Generation} RAR~\citep{yu2024randomized} modifies standard NTP training by randomly reordering the raster-order sequence and then linearly lowers the reordering probability, guiding training back to raster order NTP. RandAR~\citep{pang2025randar} removes the raster-order sequencing of decoder-only models and generates images via random-order next-token prediction, both during training and inference. SAR~\citep{liu2024customize} proposes next-set generation, which is obtained by splitting the sequence into arbitrary sets of multiple tokens. ii) \textbf{Modified Training Paradigms} VAR~\citep{tian2024visualautoregressivemodelingscalable} trains a multi-scale tokenizer and generates images by autoregressively producing next-resolution token maps, progressing from coarse to fine resolutions with parallel decoding within each scale step. NPP~\citep{pang2025patchpredictionautoregressivevisual} starts training with large static patches and gradually transitions to standard NTP, and does not use patches at inference time, unlike our method. While these approaches focus on improving the fidelity of AR Image Generation, our work addresses the scalability challenge posed by quadratic token growth in 2D tokenizers with increasing resolution, while maintaining generation quality. Further, our approach is orthogonal and complementary to these works and we leave this exploration to future work.

\paragraph{Token Reduction in NLP} Tokenization of text forms a fundamental part of the pipeline for processing text data used in training models for processing natural language. Despite known limitations~\citep{bostrom2020byte,saleva2023changes}, algorithms utilizing count-based merging of text units called subwords are present in the majority of Large Language Models (LLMs) today~\citep{singh2024tokenization}. The most common of these are the Byte Pair Encoding (BPE)~\citep{sennrich2016neural} tokenizer, which starts by treating each word as a sequence of characters and iteratively merges the most frequent adjacent pair to create larger tokens. Recently, Byte Latent Transformer (BLT)~\citep{pagnoni2024bytelatenttransformerpatches} introduced a dynamic byte-merging strategy based on next-byte prediction entropy from a lightweight autoregressive model, effectively reducing the token count for efficient language modeling from bytes. Our work draws inspiration from BLT, but unlike BLT, which tokenizes text bytes, we demonstrate that next-token prediction entropy can be applied at the VQ-VAE token level to merge low-information spatial tokens into larger patches for efficient AR image generation.

\paragraph{Token Reduction for Image Generation} Recent works have explored 1D tokenizers, which aim for better compression ratios and fewer tokens at the expense of 2D spatial structure: CAT~\citep{shen2025cat} primarily focuses on continuous latent space and adjusts token counts from a fixed set of compression ratios (typically 3) predicted by LLM based on the image complexity. TiTok~\citep{yu2024image} directly compresses images into 1D latent sequences and shows successful representation of 256×256 images with just 32 tokens. One-D-Piece~\citep{miwa2025onedpieceimagetokenizermeets} extends TiTok and proposes tail-token drop to concentrate information at the head of the sequence, allowing for variable length representation. Adaptive Length Image Tokenization~\citep{duggal2024adaptive} recurrently distills 2D discrete tokens into 1D tokens, with each iteration adding more tokens and representation capability. While 1D tokenizers result in fewer tokens, they are often less favored due to the loss of spatial structure, which is crucial for zero-shot editing capabilities such as outpainting and inpainting. Another line of work~\citep{ma2025token,havtorn2023msvit} focuses on token-merging techniques to reduce the token count: Token-Shuffle~\citep{ma2025token} leverages the low dimensionality of visual codes to merge local tokens along the channel dimension, effectively reducing the overall token count. However, these methods rely on fixed-scale merging, which can combine high-information regions and result in the loss of fine details during generation. In contrast, our approach combines strengths of both directions: it supports variable-length representations, as in 1D tokenizers, based on image information content, while retaining the 2D spatial structure required for zero-shot editing tasks.

\begin{algorithm}[!t]
	\caption{\algoname{} Training Algorithm}
	\label{alg:dynamic_patchification}
	\begin{algorithmic}[1]
        \small
		\Require Image tokens $\Itok = [x_0, \ldots, x_{T-1}]$, condition $C$, entropy model $\mathcal{E}_\phi$, threshold $\Hth$, max patch length $\Pmax$
		\Ensure Cross-entropy loss $\mathcal{L}_{\mathrm{CE}}$
		\State $\Ipatch \gets \textsc{Patchify}(I_{\mathrm{tok},\, 0:T-2},\, \mathcal{E}_\phi,\, \Hth,\, \Pmax)$
		\State $\Htokseq,\, \Hpatchseq \gets \textsc{Encoder}(I_{\mathrm{tok},\, 0:T-2},\, \Ipatch)$
		\State $\Hpatchpred \gets \textsc{GPT}\!\left(C ,\, \Hpatchseq\right)$ \Comment{AR on patches}
		\State $\Ptokpred \gets \textsc{Decoder}(\Htokseq, \, \Hpatchpred)$
		\State $\mathcal{L}_{\mathrm{CE}} \gets \textsc{CrossEntropyLoss}(\Ptokpred,\, \Itok)$
		\State \Return $\mathcal{L}_{\mathrm{CE}}$
	\end{algorithmic}
\end{algorithm}

\begin{algorithm}[!t]
	\caption{\algoname{} Inference Algorithm}
	\label{alg:dynamic_patchification_inference}
	\begin{algorithmic}[1]
        \small
		\Require Condition $C$, entropy model $\mathcal{E}_\phi$, threshold $\Hth$, max patch length $\Pmax$, target token length $T$
		\Ensure Generated image tokens $\Itok$
		\State $\Itok \gets [\,]$
		\For{$t = 0$ to $T-1$}
		\State $\Ipatch \gets \textsc{Patchify}(\Itok,\, \mathcal{E}_\phi,\, \Hth,\, \Pmax)$
		\State $e_t \gets \mathcal{H}(x_{<t};\, \mathcal{E}_\phi)$
		\If{$e_{t} \leq \Hth$ \textbf{and} $|P_m| < \Pmax$}
		\State $\Htokseq,$ \textbf{\_} $\gets \textsc{Encoder}(\Itok,\, \Ipatch)$
		\Else
		\State $\Htokseq,\, \Hpatchseq \gets \textsc{Encoder}(\Itok,\, \Ipatch)$
		\State $\Hpatchpred \gets \textsc{GPT}\!\left(C ,\, \Hpatchseq\right)$ \Comment{next-patch}
		\EndIf
		\State $\Ptokpred \gets \textsc{Decoder}(\Htokseq, \, \Hpatchpred)$
		\State $x_t \sim \hat{p}_t$ \Comment{$\hat{p}_t$ is the $t$-th distribution in $\Ptokpred$}
		\State $\Itok \gets \Itok \cup [x_t]$
		\EndFor
		\State \Return $\Itok$
	\end{algorithmic}
\end{algorithm}

%% file: sec/3_methodology.tex
\section{Methodology}

\subsection{Preliminaries}

\paragraph{Autoregressive Image Generation} Decoder-only autoregressive models represent an image $I \in \mathbb{R}^{H \times W \times 3}$ as a sequence of discrete 1D tokens and predict each token sequentially, conditioned on the previous tokens. Formally, given an image-condition pair $(I, C) \in \mathcal{D}$ and its tokenized representation $\Itok = [x_0,\, x_1,\, \ldots,\, x_{T-1}],$ where $x_i \in \{0, 1, \ldots, V-1\}$ is the $i$-th token in an ordered sequence from a vocabulary of size $V$ and $T$ is the total number of tokens. The model learns the likelihood of $I$ as a product of autoregressive factors:
\begin{equation}
    \label{eq:ar_model}
 P(\Itok \mid C) = \prod_{t=0}^{T-1} P_{\theta}(x_t \mid C, x_{<t}),
\end{equation}
where $C$ denotes an optional conditioning signal (e.g., class label or text prompt), and $P_{\theta}$ is parameterized by a decoder-only transformer with parameters~$\theta$. The model is trained to minimize the cross-entropy loss between the predicted next-token probabilities $\Ptokpred = [\hat{p}_0, \ldots, \hat{p}_{T-1}]$, where
$\hat{p}_t = P_\theta(\cdot \mid C, x_{<t})$ and the ground-truth tokens $\Itok$.
\begin{equation}
    \label{eq:ce_loss}
 \mathcal{L}_{\mathrm{CE}}
 = - \sum_{t=0}^{T-1} \log \hat{p}_t(x_t).
\end{equation}
During inference, the model generates images by sampling tokens sequentially from the learned distribution until the token length $T$ is reached.

\paragraph{Tokenization} One popular choice for image tokenization is VQ-VAE which encodes images into discrete 2D grid of tokens. Formally, given an image $I \in \mathbb{R}^{H \times W \times 3}$, VQ-VAE downsamples images by a factor of $K$ and maps each latent to a discrete token $x_i$ in the codebook of size $V$, i.e $x_i \in \{0, 1, \ldots, V-1\}$, resulting in a total of $T = \lceil\frac{H}{K}\rceil \times \lceil\frac{W}{K}\rceil$ tokens. The tokenized 1D sequence is obtained by raster-scanning the 2D tokens from top-left to bottom-right. While this approach has shown promising results in generating high-fidelity images, it suffers from scalability issues as the number of tokens $T$ scales quadratically with image resolution. For instance, increasing the image resolution from $256$px to $1024$px results in a $16\times$ increase in token count, substantially increasing the computational and memory demands of attention~\citep{tay2022efficient,dao2022flashattention,zhai2022scaling} in decoder-only AR models. \algoname{} aims to address a quadratic increase in token count by dynamically merging low-information tokens into patches guided by next-token prediction entropy, using simple encoder-decoder modules, resulting in efficient AR image generation.

\input{tables/model_description.tex}

\subsection{Dynamic Patchification for Efficient Autoregressive Image Generation}

\textbf{\algoname{}} dynamically aggregates discrete 1D token sequences into a variable length \emph{patch} sequence based on image information content for efficient generation. Our method comprises four main components (\Cref{fig:methodology}): (1) a lightweight entropy model that computes next-token prediction entropy for tokens; (2) a patch encoder that aggregates tokens within the same patch into a patch representation of the same dimensionality as each token; (3) a decoder-only transformer that operates on patch representations for efficient autoregressive generation; and (4) a patch decoder that reconstructs individual tokens from the generated patches. We use uppercase symbols to denote sequence-level quantities and hyperparameters, and lowercase symbols to denote individual token or patch states and scalar values.

\subsubsection{Patchification}
\label{sec:patchification}

The goal of patchification is to assign each token in the 1D sequence $\Itok = [x_0, \ldots, x_{T-1}]$ to a patch index such that all token indices within a patch remain contiguous. Formally, we construct a patch sequence $\Ipatch = [P_0, \ldots, P_{M-1}]$, where each patch $P_m = [s_m, \ldots, f_m]$ is a contiguous non-overlapping span of token indices starting at $s_m$ and ending at $f_m$ in the original token sequence. The number of patches $M$ varies per image and is typically smaller than the total token count $T$.

Inspired by BLT~\citep{pagnoni2024bytelatenttransformerpatches}, we propose to use next-token prediction entropy as a measure of information content for patch formation. We train a lightweight \emph{unconditional} GPT-style AR model with $C = \varnothing$ following Eqs. (\ref{eq:ar_model}) and (\ref{eq:ce_loss}) and refer to this model as the \emph{entropy model} $\mathcal{E}_\phi$ parameterized by $\phi$. We set the next-token prediction entropy for first token, $e_0 = +\infty$ to ensure it starts a new patch. The next-token prediction entropy for the token $i \in [1, T-1]$ is computed as:
\begin{equation}
	\begin{aligned}
		e_i
		 & = \mathcal{H}(x_{<i};\, \mathcal{E}_\phi) \\[3pt]
		 & = - \sum_{c=0}^{V-1}
		\mathcal{E}_\phi(x_i = c \mid x_{<i})
		\log \mathcal{E}_\phi(x_i = c \mid x_{<i})\,.
	\end{aligned}
\end{equation}
and $\Etokseq = [e_0, e_1, \ldots, e_{T-1}] = \mathcal{H}(\Itok;\, \mathcal{E}_\phi)$ denotes the entropy values for all tokens in the image. We add a token $x_i$ to the current patch $P_m$ if $e_i \le \Hth$, and start a new patch $P_{m+1}$ when $e_i > \Hth$. We further limit the patch to a maximum length of $\Pmax$ tokens to prevent excessive aggregation that could lead to information loss, and start a new patch at the end of each row to account for discontinuities in image features at row boundaries. $\Hth$ and $\Pmax$ are hyperparameters chosen via ablation, and together they determine the average patch length $\Pavg = \mathbb{E}[T/M]$ for a given dataset.

\subsubsection{Patch Encoder: Tokens to Patches}
\label{sec:patch_encoder}

The encoder is a lightweight module that encodes tokens within each patch into a patch representation and builds upon the BLT~\citep{pagnoni2024bytelatenttransformerpatches} local encoder architecture. Each patch encoder block consists of causal self-attention (SA) among tokens with 2D Rotary Positional Embedding (RoPE)~\citep{su2023roformerenhancedtransformerrotary} for spatial positional encoding:
\begin{equation}
	\Htokseq
	= \bigl[h_{x_0}, \ldots, h_{x_{T-1}}\bigr]
	= \SA\!\bigl(\Ztokseq\bigr)
\end{equation}
where $\Ztokseq = \bigl[z_{x_0}, \ldots, z_{x_{T-1}}\bigr]$, $z_{x_i}$ is the embedding of token $x_i$, and $h_{x_i}$ is its latent representation after self-attention. This is followed by cross-attention (CA)~\citep{transformer} with patches as query and tokens as keys/values. Let $\Zpatchseq = \bigl[z_{P_0}, \ldots, z_{P_{M-1}}\bigr]$ denote the initial patch queries, and let $\Htokseq_{s_m:f_m} = \bigl[h_{x_{s_m}}, \ldots, h_{x_{f_m}}\bigr]$ denote the token span corresponding to patch $P_m$. Then,
\begin{equation}
	h_{P_m} = \CA\!\bigl(z_{P_m},\, \Htokseq_{s_m:f_m}\bigr), \quad \forall\, P_m \in \Ipatch
\end{equation}
and the resulting patch representations are $\Hpatchseq = \bigl[h_{P_0}, h_{P_1}, \ldots, h_{P_{M-1}}\bigr]$.

\subsubsection{Patch Transformer}
\label{sec:patch_transformer}

The patch transformer is a decoder-only model that operates on patch representations conditioned on a class label or prompt token~$C$. By processing patches instead of individual tokens, it substantially reduces the computational cost of attention while remaining the most compute-intensive component of the pipeline. Following LlamaGen~\citep{sun2024autoregressive}, we adopt the LLaMA~\citep{llama1} architecture as our autoregressive decoder backbone and follows typical transformer design with causal self-attention and MLP layers. Given $\Hpatchseq$, the patch transformer predicts the next-patch representations $\Hpatchpred$. However, unlike standard transformers that operate on tokens and utilize 2D RoPE for positional encoding, our patch transformer employs a Dynamic RoPE mechanism (see Appendix) for encoding patches with variable token lengths.

\subsubsection{Patch Decoder: Patches to Tokens}
\label{sec:patch_decoder}

The decoder is a lightweight module that maps patches back to individual tokens, and is inspired by the local decoder architecture in BLT~\citep{pagnoni2024bytelatenttransformerpatches}. Given predicted patch states $\Hpatchpred = \bigl[\hat{h}_{P_0}, \ldots, \hat{h}_{P_{M-1}}\bigr]$, each decoder block consists of a copy operation where patches are copied to their corresponding tokens followed by norm and linear projection:
\begin{equation}
	\tilde{h}_{x_i} = h_{x_i} + \LinearOp\!\bigl(\NormOp(\hat{h}_{P_m})\bigr), \quad \forall\, i \in P_m
\end{equation}
Let $\Htokmid = \bigl[\tilde{h}_{x_0}, \ldots, \tilde{h}_{x_{T-1}}\bigr]$ denote the token states after this update. We then apply causal self-attention among tokens with 2D RoPE for positional encoding:
\begin{equation}
	\Htokout
	= \bigl[\bar{h}_{x_0}, \ldots, \bar{h}_{x_{T-1}}\bigr]
	= \SA\!\bigl(\Htokmid\bigr)
\end{equation}
Finally, a linear output head maps $\Htokout$ to the next-token probability sequence $\Ptokpred$ used for training and sampling.

\input{images/res_scaling.tex}

%% file: tables/model_description.tex
\begin{table}[t]
	\centering
	\resizebox{\linewidth}{!}{
		\footnotesize
		\setlength{\tabcolsep}{6pt}
		\renewcommand{\arraystretch}{1}
		\begin{tabular}{lccccccc}
			\toprule
			\multirow{2}{*}{\textbf{Model}}
			                & \multirow{2}{*}{\textbf{Params}}
			                & \multicolumn{3}{c}{\textbf{Layers}}
			                & \multirow{2}{*}{\textbf{Hidden}}
			                & \multirow{2}{*}{\textbf{Heads}}                                                                   \\
			\cmidrule(lr){3-5}
			                &                                     & \textbf{Enc.} & \textbf{Global T.} & \textbf{Dec.} &      &    \\
			\midrule
			\algoname{}-B   & 120M                                & 1             & 8               & 3             & 768  & 12 \\
			\algoname{}-L   & 352M                                & 1             & 19              & 4             & 1024 & 16 \\
			\algoname{}-XL  & 789M                                & 1             & 30              & 5             & 1280 & 20 \\
			\algoname{}-XXL & 1.4B                                & 1             & 41              & 6             & 1536 & 24 \\
			\bottomrule
		\end{tabular}
	}
	\caption{\textbf{Architectural specifications of \algoname{} model variants.} Enc., Global T., and Dec. refer to patch encoder, global transformer, and patch decoder, layers respectively.}
	\label{tab:model_description}
\end{table}

%% file: images/res_scaling.tex
\begin{figure}[t]
    \includegraphics[height=4.5cm, width=\columnwidth, keepaspectratio]{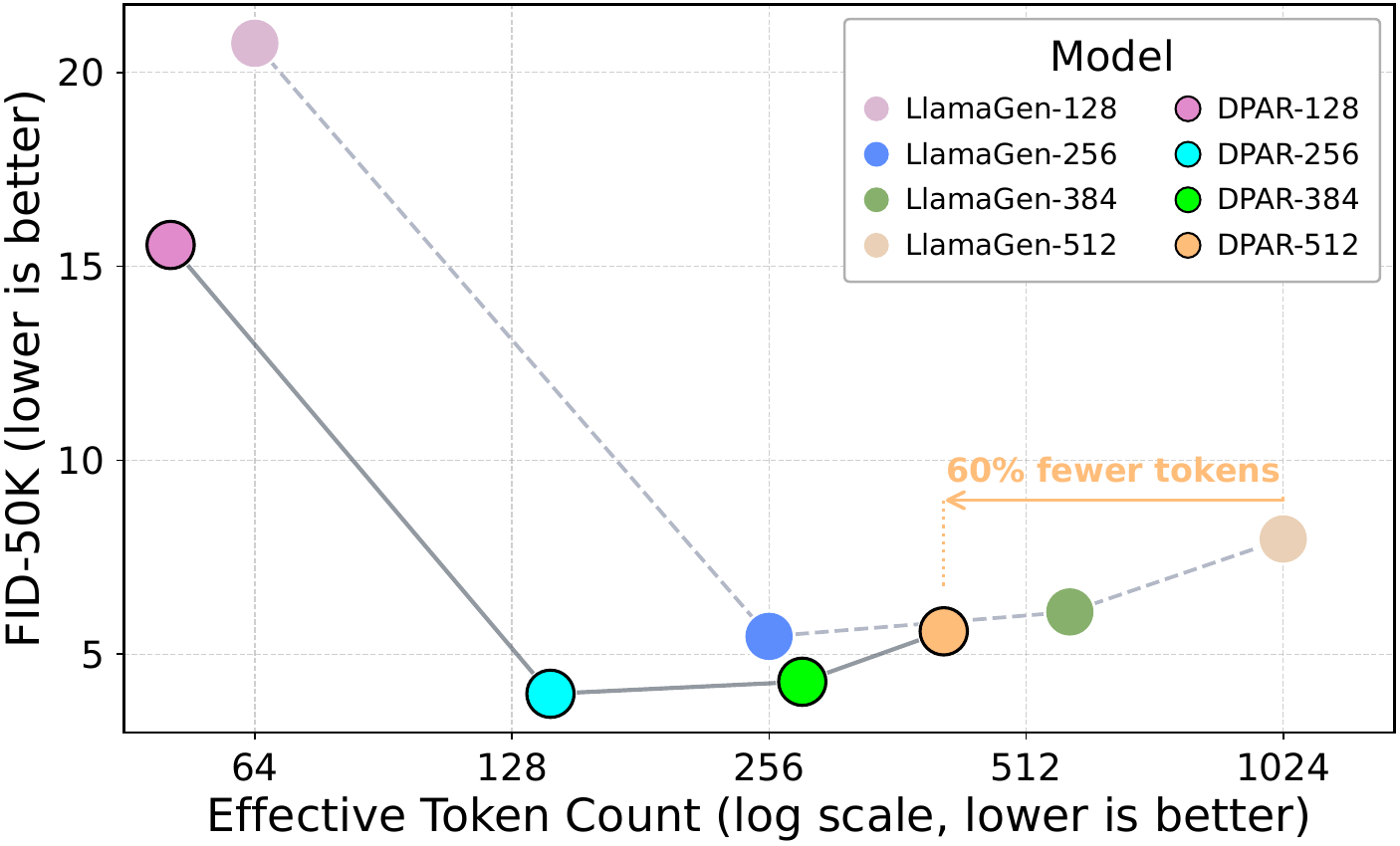}
    \caption{\textbf{\algoname{} scales more efficiently than LlamaGen across resolutions}. Comparison of \algoname{}-B and LlamaGen-B models trained at resolutions of 128, 256, 384, and 512. \algoname{} yields sub-quadratic growth in patch count with resolution, and achieves consistently lower FID at every resolution — demonstrating favorable quality-efficiency scaling.}
    \label{fig:res_scaling}
\end{figure}

%% file: sec/4_experiments.tex
\section{Experiments}

\subsection{Implementation Details}

\paragraph{Tokenizer} We use LlamaGen~\citep {sun2024autoregressive} VQ tokenizer trained on ImageNet~\citep{russakovsky2015imagenetlargescalevisual} with codebook size of $V=16384$ and downsampling of $K=16$, resulting in $T=256$ and $T=576$ tokens for 256×256 and 384×384 image respectively.

\paragraph{Entropy Model} We train an unconditional 111M LlamaGen-B on ImageNet using the same training setup as the patch transformer. For \algoname{} training with 256×256 images, we use $\Hth = 7.8$ and $\Pmax = 4$, resulting in an average patch length of $\Pavg = 1.81$ on the ImageNet training set. For 384×384 images, we use $\Hth = 7.9$ and $\Pmax = 4$, resulting in $\Pavg = 2.06$.

\paragraph{\algoname{} Architecture} We implement \algoname{} with minimal modifications on the LlamaGen architecture. Across all \algoname{} model variants, we use a single-layer encoder, as our ablations did not show any meaningful gains from deeper encoders (see Appendix). We start with 3 decoder layers for the Base model and increase the number of decoder layers at the same rate as the hidden dimensions (i.e., doubling the hidden dimension from 768 to 1536 increases the decoder layers from 3 to 6). Overall, the aim is to keep both the encoder and decoder shallow, so that the majority of the computational budget is allocated to the patch transformer, which operates on variable-length patches, thereby enabling efficient computation. Furthermore, for each variant, we set the number of patch transformer layers to ensure the total layer count matches that of LlamaGen models of comparable size for a fair comparison. Detailed \algoname{} architectural specifications are provided in ~\Cref{tab:model_description}.

\paragraph{Training and Sampling} We train class-conditional \algoname{} models on ImageNet for resolutions 256×256 and 384×384. Our models are trained on 8 A100 GPUs with a global batch size of 256 for 300 epochs (${\approx}1.5$M steps) using the AdamW~\citep{loshchilov2019decoupledweightdecayregularization} optimizer with a learning rate of 1e-4, weight decay of 0.05, $\beta_1=0.9$, $\beta_2=0.95$, and gradient clipping of 1.0 with PyTorch DDP. Other settings, including data augmentation, follow LlamaGen~\citep{sun2024autoregressive}. We do not use any learning rate scheduling and maintain a constant learning rate throughout training. We also use classifier-free guidance~\citep{ho2022classifierfreediffusionguidance} with a drop probability of 0.1 during training and preprocess tokens and entropy values for each image in the dataset to avoid on-the-fly computation during training. Furthermore, since the patch transformer operates on a variable patch sequence, we implement a packed variant of LlamaGen with xformers~\citep{xFormers2022} to efficiently process a batch without padding to the maximum sequence length. For sampling, we follow the sampling strategy of LlamaGen~\citep{sun2024autoregressive} and use top-k sampling with $k=0$, top-p of $1.0$, and temperature of $1.0$ for all our experiments.

\input{tables/main_results.tex}

\input{images/fid_epochs.tex}

\subsection{ImageNet-256 Benchmark}

We evaluate the performance of our method on the class-conditional ImageNet-256 generation benchmark, a popular benchmark for image generation~\citep{sun2024autoregressive,pang2025randar, peebles2023scalablediffusionmodelstransformers}. We train class-conditional \algoname{} model variants on ImageNet at resolutions 256×256 and 384×384 and measure model performance primarily with FID-50K~\citep{heusel2018ganstrainedtimescaleupdate}, which computes FID between 50,000 generated samples and the ImageNet validation dataset. For 384×384 resolution, the generated images are resized to 256×256 before computing FID, as in prior work~\citep{sun2024autoregressive,pang2025randar}. We also report Inception Score (IS)~\citep{salimans2016improvedtechniquestraininggans}, and precision/recall~\citep{kynkaanniemi2019improved} for completeness. We primarily compare our model with LlamaGen, as other works either utilize different tokenizers~\citep{tian2024visual}, token orderings~\citep{pang2025randar,yu2024randomized,liu2024customize}, and training paradigms, making direct comparisons unfair. We leave the extension of our method to these approaches for future work. Our results are summarized in~\Cref{tab:main_results}. We observe that \algoname{} outperforms LlamaGen at both resolutions across all model sizes, improving the FID score by up to 29.6\% on the Base model at 384×384 resolution. We further compare the convergence speed of \algoname{} and LlamaGen by plotting FID against training epochs in~\Cref{fig:fid_epochs} for models trained on resolution 384×384 due to the availability of intermediate epoch results from LlamaGen. We observe that \algoname{} consistently outperforms LlamaGen throughout training, demonstrating faster convergence and better final performance. Finally, we study the scaling behavior of \algoname{}-B across resolutions 128, 256, 384, and 512. As shown in~\Cref{fig:res_scaling}, \algoname{} attains lower FID than LlamaGen while using up to 60\% fewer tokens, highlighting better quality-efficiency scaling.

\subsection{Training FLOPs}

We estimate the mean training cost by measuring FLOPs over 500 training steps and averaging the total FLOP measurements across the entire run to obtain a per-sample FLOP estimate. This ensures a consistent, profile-based estimate of the end-to-end compute required by \algoname{} with variable-length patches. The results are summarized in Fig.~\ref{fig:teaser}(b). We observe that \algoname{} significantly reduces training FLOPs across all model sizes compared to LlamaGen, primarily due to its packed patch transformer implementation, which efficiently handles variable-length patches without padding. Notably, \algoname{}-XL achieves a 40.4\% reduction in training FLOPs compared to LlamaGen-XL at a 384px generation resolution on ImageNet.

\subsection{Ablation Studies}

We use the \algoname{}-L model trained on ImageNet at 256px resolution for 50 epochs for ablations, unless specified otherwise, and compare  ablation choices using FID-50K.

\paragraph{Patchification Strategies} We consider three binary design choices for patchification: i) starting a new patch when the entropy exceeds a threshold $\Hth$ (entropy gating), ii) limiting the maximum patch length to $\Pmax$ (maximum patch length), and iii) resetting patches at row boundaries (row-boundary resets). When all three design choices are disabled, this corresponds to the static patchification scheme with fixed patch length $\Pstatic$, and we set $\Pstatic = 1.81$ to match the average patch length of our full method. The results are summarized in ~\Cref{tab:tokenization_strategies}. We observe that enabling all three design choices leads to the best FID score of 3.32. Furthermore, entropy alone is insufficient to achieve optimal performance, as the absence of a limit on the maximum patch length can lead to excessive aggregation, potentially resulting in information loss.

\begin{table}[t]
	\centering
	\setlength{\tabcolsep}{4pt}
	\renewcommand{\arraystretch}{0.95}
	\scriptsize
	\begin{tabular}{@{}c@{\hskip 3pt}c@{\hskip 3pt}cc@{}}
		\toprule
		\textbf{Entropy Gating} & \textbf{Max Patch Len.} & \textbf{Row-Bound. Reset} & \textbf{FID} $\downarrow$ \\
		\midrule
		$\times$      & $\times$            & $\times$        & 3.58                      \\
		\checkmark    & $\times$            & $\times$        & 3.91                      \\
		\checkmark    & \checkmark          & $\times$        & 3.45                      \\
		\checkmark    & \checkmark          & \checkmark      & \textbf{3.32}             \\
		\bottomrule
	\end{tabular}
	\caption{\textbf{Patchification strategies}. Entropy-based patchification with max patch length and row-boundary reset constraint leads to the best FID score on ImageNet 256×256 benchmark.}
	\label{tab:tokenization_strategies}
\end{table}

\paragraph{Effect on Entropy Threshold $\Hth$}

We study the impact of the entropy threshold $\Hth$ on FID with a fixed maximum patch length $\Pmax = 4$ (see \Cref{tab:entropy_ablations}). Lower thresholds produce smaller patches and increase computational cost, whereas higher thresholds produce larger patches but degrade image quality. An intermediate value of $\Hth = 7.8$ provides the best balance, achieving the lowest FID of 3.32.

\paragraph{Effect of Max Patch Length $\Pmax$}

We examine the effect of the maximum patch length $\Pmax$ on FID with fixed entropy threshold $\Hth = 7.8$ (see \Cref{tab:entropy_ablations}). We observe that FID first improves as we increase $\Pmax$ from 1 to 4, achieving the best FID of 3.32 at $\Pmax = 4$. However, further increasing $\Pmax$ results in performance degradation, likely due to information loss from excessive aggregation.

\input{tables/entropy_threshold.tex}

\subsection{Adaptive Patch Lengths at Inference}
We investigate whether \algoname{} models trained with a fixed entropy threshold $\Hthtrain$ can generalize to different inference thresholds $\Hthinf$ at inference time. To evaluate this, we take our \algoname{}-L model trained with $\Hthtrain = 7.8$ and $\Pmax = 4$ for 50 epochs and vary $\Hthinf$ during inference from 7.8 to 8.1, and compare with a static model trained with fixed patch length $\Pstatic = 1.81$ in the ablation studies. The results are summarized in ~\Cref{tab:threshold_extrapolation}. We observe that as $\Hthinf$ increases from 7.8 to 8.1, \algoname{} FID degrades marginally from 3.32 to 3.39, while the static model's FID degrades significantly, from 3.58 to 25.59.

We argue that \algoname{} with dynamic patchification yields stronger global representations, as a patch must keep track of future tokens across variable-length sequences. This enables better generalization, resulting in adaptive patch sizes at inference without significant degradation in FID. We further test our hypothesis by linearly probing the global image feature obtained by averaging the patch/token representations of the last hidden layer of the \algoname{}-L patch transformer and the LlamaGen-L transformer. \algoname{}-L achieves 37.82\% top-1 and 61.15\% top-5 accuracy versus 32.62\% and 56.61\% for LlamaGen-L — a $\sim$5 pp. improvement in both metrics — suggesting that \algoname{} learns stronger global representations.

\input{tables/tokenization_strategies.tex}

%% file: tables/main_results.tex
\begin{table*}[t]
	\centering
	\label{table:main_results}
	\resizebox{0.7\linewidth}{!}{
		\footnotesize
		\setlength{\tabcolsep}{6pt}
		\renewcommand{\arraystretch}{1.05}
		\begin{tabular}{l l c c c c c c}
			\toprule
			\textbf{Type}       & \textbf{Model}                                & \textbf{\#Params} & \textbf{FID↓} & \textbf{IS↑} & \textbf{Prec.↑} & \textbf{Rec.↑} & \textbf{Steps} \\
			\midrule
			Diffusion
			                    & ADM~\citep{dhariwal2021diffusion}             & 554M              & 4.59          & 186.70       & 0.82            & 0.523          & 250            \\
			                    & LDM-4~\citep{rombach2022high}                 & 400M              & 3.60          & 247.70       & --              & --             & 250            \\
			                    & DiT-XL~\citep{peebles2023scalable}            & 675M              & 2.27          & 278.20       & 0.83            & 0.57           & 250            \\
			                    & SiT-XL~\citep{ma2024sit}                      & 675M              & 2.06          & 270.30       & 0.82            & 0.59           & 250            \\
			\midrule
			Bidirectional AR
			                    & MaskGIT-re~\citep{chang2022maskgit}           & 227M              & 4.02          & 355.60       & --              & --             & 8              \\
			                    & MAGVIT-v2~\citep{yu2023language}              & 307M              & 1.78          & 319.40       & --              & --             & 64             \\
			                    & MAR-H~\citep{li2024autoregressive}            & 943M              & 1.55          & 303.70       & 0.81            & 0.62           & 256            \\
			                    & TiTok-S-128~\citep{yu2024image}               & 287M              & 1.97          & 281.80       & --              & --             & 64             \\
			\midrule
			Causal AR           & VAR~\citep{tian2024visual}                    & 600M              & 2.57          & 302.60       & 0.83            & 0.56           & 10             \\
			(non-raster order / & VAR~\citep{tian2024visual}                    & 2.0B              & 1.92          & 350.20       & 0.82            & 0.59           & 10             \\
			modified-training)  & SAR-XL~\citep{liu2024customize}               & 893M              & 2.76          & 273.80       & 0.84            & 0.55           & 256            \\
			                    & RAR-L~\citep{yu2024randomized}                & 461M              & 1.70          & 299.50       & 0.81            & 0.60           & 256            \\
			                    & RAR-XXL~\citep{yu2024randomized}              & 955M              & 1.50          & 306.90       & 0.80            & 0.62           & 256            \\
			                    & RAR-XL~\citep{yu2024randomized}               & 1.5B              & 1.48          & 326.00       & 0.80            & 0.63           & 256            \\
			                    & RandAR-L~\citep{pang2025randar}               & 343M              & 2.55          & 288.82       & 0.81            & 0.58           & 88             \\
			                    & RandAR-XL~\citep{pang2025randar}              & 775M              & 2.25          & 317.77       & 0.80            & 0.60           & 88             \\
			                    & RandAR-XL~\citep{pang2025randar}              & 775M              & 2.22          & 314.21       & 0.80            & 0.60           & 256            \\
			                    & RandAR-XXL~\citep{pang2025randar}             & 1.4B              & 2.15          & 321.97       & 0.79            & 0.62           & 88             \\
			\midrule
			Causal AR           & VQGAN-re~\citep{esser2021taming}              & 1.4B              & 5.20          & 280.30       & --              & --             & 256            \\
			(raster order)      & RQTran.-re~\citep{lee2022autoregressive}      & 3.8B              & 3.80          & 323.70       & --              & --             & 64             \\
			                    & Open-MAGVIT2-XL~\citep{luo2024open}           & 1.5B              & 2.33          & 271.80       & 0.84            & 0.54           & 256            \\
			\midrule
			Causal AR           & LlamaGen-B~\citep{sun2024autoregressive}      & 111M              & 5.46          & 193.61       & 0.84            & 0.46           & 256            \\
			(raster order       & LlamaGen-L~\citep{sun2024autoregressive}      & 343M              & 3.80          & 248.30       & 0.83            & 0.52           & 256            \\
			with LlamaGen       & LlamaGen-XL~\citep{sun2024autoregressive}     & 775M              & 3.39          & 227.10       & 0.81            & 0.54           & 256            \\
			tokenizer)          & LlamaGen-384-B~\citep{sun2024autoregressive}  & 111M              & 6.09          & 182.53       & 0.84            & 0.42           & 576            \\
			                    & LlamaGen-384-L~\citep{sun2024autoregressive}  & 343M              & 3.07          & 256.06       & 0.83            & 0.52           & 576            \\
			                    & LlamaGen-384-XL~\citep{sun2024autoregressive} & 775M              & 2.62          & 244.08       & 0.80            & 0.57           & 576            \\
								& LlamaGen-384-XXL~\citep{sun2024autoregressive} & 1.4B              & 2.34       & 253.91             & 0.81             & 0.60             & 576            \\
			                    & \algoname{}-B (cfg=2.1)                       & 120M              & 3.98          & 250.62       & 0.83            & 0.49           & 142            \\
			                    & \algoname{}-L (cfg=1.9)                       & 352M              & 2.93          & 269.34       & 0.81            & 0.56 & 142            \\
			                    & \algoname{}-XL (cfg=2.0)                      & 789M              & 2.67          & 281.65       & 0.82            & 0.56              & 142            \\
			                    & \algoname{}-384-B (cfg=2.10)                   & 120M              & 4.29          & 254.54       & 0.83            & 0.47 & 280            \\
			                    & \algoname{}-384-L (cfg=1.90)                   & 352M              & 2.79          & 283.84       & 0.81            & 0.55 & 280            \\
			                    & \algoname{}-384-XL (cfg=1.90)                  & 789M              & 2.60          & 285.43       & 0.81            & 0.57              & 280            \\
								& \algoname{}-384-XXL (cfg=1.75)                  & 1.4B              & 2.30       & 287.37             & 0.82             & 0.59             & 280            \\
			\bottomrule
		\end{tabular}
	}
    \caption{\textbf{\algoname{} model comparisons on class-conditional ImageNet 256×256 benchmark.} We report FID~\citep{heusel2018ganstrainedtimescaleupdate}, Inception Score(IS)~\citep{salimans2016improvedtechniquestraininggans}, and precision/recall ~\citep{kynkaanniemi2019improved} and the average number of sampling steps used for generation. \algoname{} model outperforms prior raster-order autoregressive models with similar parameter counts, achieving significantly better FID scores. Models containing `-384' in their names are trained on $384\times384$ and resized to $256\times256$ for evaluation.}
	\label{tab:main_results}
\end{table*}

%% file: images/fid_epochs.tex
\begin{figure}[t]
    \includegraphics[height=4.5cm, width=\columnwidth, keepaspectratio]{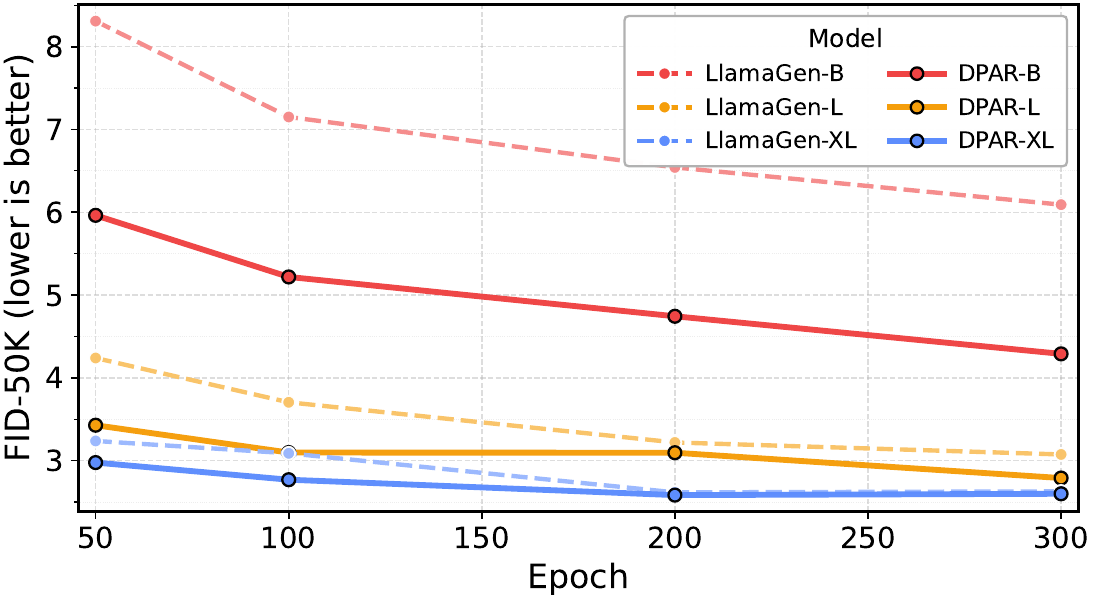}
    \caption{\textbf{\algoname{} converges faster than LlamaGen across model sizes.} FID vs.\ training epochs on ImageNet-384 for B, L, and XL variants. \algoname{} consistently achieves lower FID throughout training — demonstrating faster convergence and quality.}
    \label{fig:fid_epochs}
\end{figure}

%% file: tables/entropy_threshold.tex
\begin{table}[t]
    \centering
    \footnotesize
    \setlength{\tabcolsep}{6pt}
    \renewcommand{\arraystretch}{1.05}
    \begin{subtable}{\linewidth}
        \centering
        \begin{tabular}{lccccc}
            \toprule
            \textbf{$\Hth$}          & \textbf{7.6} & \textbf{7.7} & \textbf{7.8}  & \textbf{7.9} & \textbf{8.0} \\
            \midrule
            \textbf{FID-50K ($\downarrow$)} & 3.36         & 3.41         & \textbf{3.32} & 3.46         & 3.43         \\
            \bottomrule
        \end{tabular}
        \caption{Varying $\Hth$ with $\Pmax = 4$ fixed.}
        \label{tab:entropy_thresholds}
    \end{subtable}
    \begin{subtable}{\linewidth}
        \centering
        \begin{tabular}{lcccccc}
            \toprule
            \textbf{$\Pmax$}          & \textbf{1} & \textbf{2} & \textbf{4}    & \textbf{6} & \textbf{8} & \textbf{16} \\
            \midrule
            \textbf{FID-50K ($\downarrow$)} & 3.73       & 3.42       & \textbf{3.32} & 3.49       & 3.41       & 3.61        \\
            \bottomrule
        \end{tabular}
        \caption{Varying $\Pmax$ with $\Hth = 7.8$ fixed.}
        \label{tab:entropy_max_plen}
    \end{subtable}
    \caption{\textbf{Ablation on entropy threshold $\Hth$ and maximum patch length $\Pmax$.} $\Hth = 7.8$ and $\Pmax = 4$ jointly achieve the best FID of 3.32.}
    \label{tab:entropy_ablations}
\end{table}

%% file: tables/tokenization_strategies.tex
\begin{table}[t]
	\centering
	\setlength{\tabcolsep}{4pt}
	\renewcommand{\arraystretch}{0.95}
	\scriptsize
	\begin{tabular}{lcccc}
		\toprule
		\textbf{$\Hthinf$}  & \textbf{7.8}  & \textbf{7.9}  & \textbf{8.0}  & \textbf{8.1}  \\
		\midrule
		\textbf{$\Pavg$} & 1.81          & 1.92          & 2.03          & 2.16          \\
		\midrule
		Static                      & 3.58          & 7.85          & 17.91         & 25.59         \\
		\algoname{}-L               & 3.31          & 3.32          & 3.38          & 3.39          \\
		\bottomrule
	\end{tabular}
	\caption{\textbf{Adaptive Patch Length at Inference.} \algoname{}-L trained with $\Hthtrain{=}7.8$ maintains stable FID as $\Hthinf$ increases from 7.8 to 8.1, whereas the static model degrades substantially.}
	\label{tab:threshold_extrapolation}
\end{table}

%% file: sec/5_conclusion.tex
\section{Conclusion}

In this work, we introduced \algoname{}, a novel autoregressive image generation model that leverages next-token prediction entropy to dynamically aggregate tokens into variable-length patches, reducing token count and training FLOPs for efficient image generation. While this work focuses on raster-order generation, our proposed strategy is compatible with random-order methods~\citep{pang2025randar,yu2024randomized}, and extending it to these settings as well as developing KV-Cache strategies tailored to the encoder-decoder architecture for inference speedups are promising directions for future work. 

%% file: sec/X_supplementary.tex
\clearpage
\hypersetup{pageanchor=false}
\setcounter{table}{0}
\setcounter{figure}{0}
\setcounter{equation}{0}
\setcounter{algorithm}{0}
\renewcommand{\wrap}[1]{#1}
\setcounter{section}{0}
\renewcommand{\thetable}{S\arabic{table}}
\renewcommand{\thefigure}{S\arabic{figure}}
\renewcommand{\theequation}{S\arabic{equation}}
\maketitlesupplementary
\hypersetup{pageanchor=true}

\section{Training}\label{sec:appendix_scaling_factor}

\begin{algorithm}[th]
    \caption{\algoname{} Patchification Algorithm}
    \label{alg:patchification}
    \begin{algorithmic}[1]
        \Require Image tokens $\Itok = [x_0, x_1, \ldots, x_{T-1}]$, entropy model $\mathcal{E}_\phi$, threshold $\Hth$, max patch length $\Pmax$
        \Ensure Patch sequence $\Ipatch$
        \State $\Ipatch \gets [[x_0]]$ \Comment{Initialize first patch with first token}
        \State $P \gets [x_1]$ \Comment{Current patch}
        \For{$i = 2$ to $T-1$}
        \State $e_{i} \gets \mathcal{H}(x_{<i};\, \mathcal{E}_\phi)$
        \If{$e_{i} \leq \Hth$ \textbf{and} $|P| < \Pmax$ \textbf{and} $x_i$ not at row start}
        \State $P \gets P \cup [x_{i}]$ \Comment{Add token to current patch}
        \Else
        \State $\Ipatch \gets \Ipatch \cup [P]$ \Comment{Finalize current patch}
        \State $P \gets [x_{i}]$ \Comment{Start new patch}
        \EndIf
        \EndFor
        \State $\Ipatch \gets \Ipatch \cup [P]$ \Comment{Add last patch}
        \State \Return $\Ipatch$
    \end{algorithmic}
\end{algorithm}

\subsection{Detailed Results}

We have provided detailed results for all model variants across different training epochs and CFG scales in \Cref{tab:complete_results_res256} and \Cref{tab:complete_results_res384}. We also provide resolution-scaling CFG sweep results across 128, 256, 384, and 512 resolutions for DPAR-B and LlamaGen-B in \Cref{tab:complete_results_res_scaling}. Finally, we have provided a training cost comparison with LlamaGen, which was used for generating FID-FLOPs chart in \Cref{tab:gflops_all}.

\subsection{Patchification Algorithm}

The patchification algorithm assigns each token in the 1D sequence $\Itok = [x_0, \ldots, x_{T-1}]$ to a patch index such that all token indices within a patch remain contiguous. Formally, we construct a patch sequence $\Ipatch = [P_0, \ldots, P_{M-1}]$, where each patch $P_m = [s_m, \ldots, f_m]$ is a contiguous non-overlapping span of token indices starting at $s_m$ and ending at $f_m$ in the original token sequence. The number of patches $M$ varies per image and is typically smaller than the total token count $T$.

\input{tables/flops.tex}

\input{tables/encoder_decoder_layers.tex}

\subsection{Ablation Study: Encoder-Decoder Layers}
We conduct an ablation study to analyze the impact of varying the number of encoder and decoder layers on model performance, keeping the total number of encoder and decoder layers constant. As shown in \Cref{tab:ed_layers_ablation}, configurations with shallower encoders and deeper decoders (E1D4) yield the best FID scores. This suggests that allocating more capacity to the decoder is beneficial for generating high-quality images, while a lighter encoder suffices for aggregating tokens into patches.

\subsection{Dynamic RoPE}
\label{sec:dynamic_rope}

2D Rotary Positional Embedding (RoPE)~\citep{su2023roformerenhancedtransformerrotary} encodes each token's 2D spatial coordinate $(x, y)$ by rotating its query and key representations in latent space with dimensionality $d$ using sinusoidal functions of the coordinates. For a token located at 2D coordinates $(x, y)$ in the image, the positional encoding is given by:
\begin{equation}
    \label{eq:rope_2d}
    \begingroup\setlength{\arraycolsep}{3pt}
    \begin{aligned}
        \omega_i            & = 10000^{-4(i-1)/d},\quad i=1,\ldots,\tfrac{d}{4},       \\
        \mathbf{r}_x        & = [\,\sin(\omega_i x),\,\cos(\omega_i x)\,]_{i=1}^{d/4}, \\
        \mathbf{r}_y        & = [\,\sin(\omega_i y),\,\cos(\omega_i y)\,]_{i=1}^{d/4}, \\
        \mathbf{r}_{(x, y)} & = [\mathbf{r}_x,\, \mathbf{r}_y]
    \end{aligned}
    \endgroup
\end{equation}
where $\omega_i$ are the frequency terms, and the resulting $\mathbf{r}_{(x, y)}$ is used to rotate the query and key vectors to encode 2D spatial relationships. We propose Dynamic RoPE, an extension of 2D RoPE to handle variable length patches by encoding the start and end coordinates of each patch along the y-axis. This is possible since each row starts a new patch. The updated positional encoding for a patch $P_m$ that spans tokens from $(x, y_{s_m})$ to $(x, y_{f_m})$ is defined as:
\begin{equation}
    \label{eq:dynamic_rope_2d}
    \begingroup\setlength{\arraycolsep}{3pt}
    \begin{aligned}
        \omega_i                   & = 10000^{-4(i-1)/d}, \quad i = 1,\ldots,\tfrac{d}{4},                                         \\[2pt]
        \alpha_i                   & = 10000^{-16(i-1)/d}, \quad i = 1,\ldots,\tfrac{d}{16},                                       \\[3pt]
        \mathbf{r}_x               & = [\,\sin(\omega_i x),\, \cos(\omega_i x)\,]_{i=1}^{d/4},                                     \\[2pt]
        \mathbf{r}_{y_s}           & = [\,\sin(\alpha_i y_{s_m}),\, \cos(\alpha_i y_{s_m})\,]_{i=1}^{d/16},                        \\[2pt]
        \mathbf{r}_{y_f}           & = [\,\sin(\alpha_i y_{f_m}),\, \cos(\alpha_i y_{f_m})\,]_{i=1}^{d/16},                        \\[3pt]
        \mathbf{r}_{(x, y_s, y_f)} & = [\mathbf{r}_x,\, \mathbf{r}_{y_s},\, \mathbf{r}_{y_f},  \mathbf{r}_{y_f}, \mathbf{r}_{y_s}]
    \end{aligned}
    \endgroup
\end{equation}
Our idea is to encode both the starting and ending y-coordinates of each patch, allowing the model to capture the horizontal span of each patch in addition to its vertical position. Further, adding redundancy by repeating the start and end positional encodings leads to better representation as observed in \Cref{tab:embedding_comparison}.
\input{tables/position_embedding_strategies.tex}

\subsection{Entropy-Model Dataset Transferability}
We evaluate whether an entropy model trained on a different dataset can still provide useful patchification for ImageNet generation. As shown in \Cref{tab:entropy_dataset_transfer}, using an entropy model trained on COCO-17 yields similar FID to an ImageNet-trained entropy model while maintaining a strong correlation between the resulting entropy maps. This suggests that entropy models trained on reasonably diverse dataset can be generalized for entropy prediction on a different dataset.

\input{tables/entropy_dataset_transfer.tex}

\subsection{Entropy-Model Size Sensitivity}
A lightweight EM is sufficient for patchification, with only marginal gains from
increasing its size. \Cref{tab:entropy_model_size} shows a lightweight 50M entropy model performs comparably to a 110M model, indicating that patchification quality is not highly sensitive to entropy-model scale and a small model is sufficient for high-quality patchification.

\input{tables/entropy_model_size.tex}

\subsection{Scaling to 1.4B Parameters}
We also scale DPAR to the XXL regime at 384 resolution to test whether the training-efficiency gains persist at higher capacity. As shown in \Cref{tab:xxl_384_scaling_appendix}, DPAR-384-XXL improves FID over the LlamaGen baseline while reducing training FLOPs by 42.37\%.

\input{tables/xxl_384_scaling_appendix.tex}

\subsection{Relation with Image Compression Algorithms}
For such images, popular compression algorithms such as PNG, which compress based on information content, also tend to have a larger size after compression. We observed a high correlation ($p$-corr = 0.54) between the image size after compression (indicative of information content) versus number of patches in an image, suggesting that NTPE aligns well with information content of an image.

\input{tables/complete_results_res256.tex}
\input{tables/complete_results_res384.tex}
\input{tables/complete_results_res_scaling.tex}

\clearpage
\input{visualizations/visualizations.tex}

%% file: tables/flops.tex
\begin{table}[t]
    \centering
    \renewcommand{\arraystretch}{1.1}
    \scriptsize

    \resizebox{\columnwidth}{!}{%
    \begin{tabular}{lccc}
        \toprule
        \textbf{Variant} & \textbf{LlamaGen} & \textbf{\algoname{}} & \textbf{Reduction} \\
                         & \textbf{(GFLOPs)} & \textbf{(GFLOPs)}   & \textbf{(\%)} \\
        \midrule
        B-256   & 24.98  & 19.21  & 23.1 \\
        L-256   & 83.26  & 56.74  & 31.9 \\
        XL-256  & 192.69 & 125.52 & 34.9 \\
        \midrule
        B-384   & 56.21  & 40.92  & 27.2 \\
        L-384   & 187.35 & 117.92 & 37.1 \\
        XL-384  & 433.57 & 258.53 & 40.4 \\
        XXL-384 & 797.59 & 459.63 & 42.4 \\
        \bottomrule
    \end{tabular}}

    \caption{\textbf{Compute comparison across all LlamaGen and \algoname{} variants}. \algoname{} consistently reduces FLOPs across both 256×256 and 384×384 model families.}
    \label{tab:gflops_all}
\end{table}

%% file: tables/encoder_decoder_layers.tex
\begin{table}
    \centering
    \setlength{\tabcolsep}{6pt}
    \renewcommand{\arraystretch}{1.05}
    \footnotesize

    \begin{tabular}{lcccc}
        \toprule
        \textbf{Layers (E\#D\#)} & \textbf{E1D4} & \textbf{E2D3} & \textbf{E3D2} & \textbf{E4D1} \\
        \midrule
        \textbf{FID-50K ($\downarrow$)} 
            & \textbf{3.32} 
            & 3.35 
            & 3.51 
            & 3.85 \\
        \bottomrule
    \end{tabular}

    \caption{\textbf{Ablation on encoder–decoder depth.}  
    $E_iD_j$ indicates $i$ encoder layers and $j$ decoder layers.  
    Shallow encoders with deeper decoders (E1D4) provide the best FID.}
    \label{tab:ed_layers_ablation}
\end{table}

%% file: tables/position_embedding_strategies.tex
\begin{table}[t]
    \centering
    \footnotesize
    \setlength{\tabcolsep}{6pt}
    \renewcommand{\arraystretch}{1.05}
    \begin{tabular}{lc}
        \toprule
        \textbf{Method}               & \textbf{FID$\downarrow$} \\
        \midrule
        2D Embedding                       & 3.32          \\
        Dynamic Embedding w/o redundancy   & 3.42          \\
        Dynamic Embedding                  & \textbf{3.31} \\
        \bottomrule
    \end{tabular}
    \caption{\textbf{Comparison of positional embedding schemes.} Dynamic Embedding achieves the best FID on ImageNet 256×256.}
    \label{tab:embedding_comparison}
\end{table}

%% file: tables/entropy_dataset_transfer.tex
\begin{table}[H]
    \centering
    \setlength{\tabcolsep}{5pt}
    \renewcommand{\arraystretch}{1.12}
    \footnotesize
    \begin{tabular}{lcc}
        \toprule
        \textbf{Entropy Model Dataset} & \textbf{DPAR-B FID} & \textbf{$p$-corr} \\
        \midrule
        ImageNet & 3.98 & 1.00 \\
        COCO-17 (118K) & 3.96 & 0.85 \\
        \bottomrule
    \end{tabular}
    \caption{\textbf{Entropy-model dataset transferability.} DPAR-B is trained on ImageNet while entropy maps are generated by entropy models trained on different datasets.}
    \label{tab:entropy_dataset_transfer}
\end{table}

%% file: tables/entropy_model_size.tex
\begin{table}[H]
    \centering
    \setlength{\tabcolsep}{5pt}
    \renewcommand{\arraystretch}{1.12}
    \footnotesize
    \begin{tabular}{lcc}
        \toprule
        \textbf{Entropy Model Size} & \textbf{DPAR-B FID} & \textbf{$p$-corr} \\
        \midrule
        110M & 3.976 & 1.00 \\
        50M & 3.982 & 0.88 \\
        \bottomrule
    \end{tabular}
    \caption{\textbf{Effect of entropy-model size.} A lightweight entropy model yields comparable patchification quality and generation performance.}
    \label{tab:entropy_model_size}
\end{table}

%% file: tables/xxl_384_scaling_appendix.tex
\begin{table}[H]
    \centering
    \setlength{\tabcolsep}{5pt}
    \renewcommand{\arraystretch}{1.12}
    \footnotesize
    \begin{tabular}{lcc}
        \toprule
        \textbf{Model} & \textbf{FID} & \textbf{Train FLOPs} \\
        \midrule
        LlamaGen-384-XXL (1.4B) & 2.34 & 797.59 \\
        DPAR-384-XXL (1.4B) & 2.30 & 459.63 (-42.4\%) \\
        \bottomrule
    \end{tabular}
    \caption{\textbf{Scaling to 1.4B parameters at 384 resolution} from rebuttal experiments.}
    \label{tab:xxl_384_scaling_appendix}
\end{table}

%% file: tables/complete_results_res256.tex
\begin{table*}[t]
    \centering
    \resizebox{\linewidth}{!}{%
        \setlength{\tabcolsep}{10pt}
        \renewcommand{\arraystretch}{1.22}
        \footnotesize
        \begin{tabular}{lccccccc}
            \toprule
            \textbf{Model} & \textbf{Params}       & \textbf{Epoch}       & \textbf{CFG} & \textbf{FID↓} & \textbf{IS↑} & \textbf{Prec.↑} & \textbf{Rec.↑} \\
            \midrule\midrule

            \multirow{4}{*}{\textbf{B-256}}
                           & \multirow{4}{*}{120M}
                           & \multirow{4}{*}{300}
                           & 1.75                  & {5.02}                 & 193.99       & 0.78          & 0.54                                            \\
                           &                       &                      & 1.90         & 4.28          & 219.40       & 0.81            & 0.52           \\
                           &                       &                      & 2.00         & 4.07          & 235.39       & 0.82            & 0.50           \\
                           &                       &                      & 2.10         & 3.98          & 250.62       & 0.83            & 0.49           \\
            \midrule\midrule

            \multirow{4}{*}{\textbf{L-256}}
                           & \multirow{4}{*}{352M}
                           &  \multirow{4}{*}{300}
                           & 1.75                  & {3.24}                 & 241.05       & 0.79          & 0.58                                            \\
                           &                       &                      & 1.90         & 2.93          & 269.34       & 0.81            & 0.56           \\
                           &                       &                      & 2.00         & 2.96          & 284.06       & 0.82            & 0.54           \\
                           &                       &                      & 2.10         & 3.03          & 298.19       & 0.83            & 0.54           \\
            \midrule\midrule

            \multirow{5}{*}{\textbf{XL-256}}
                           & \multirow{5}{*}{789M}
                           & 200                   & 2.00                 & {2.86}         & 277.37        & 0.82         & 0.56                             \\ \cline{3-8}
                           &                       & \multirow{4}{*}{300}
                           & 1.75                  & 2.82                 & 249.57       & 0.79          & 0.60                                            \\
                           &                       &                      & 1.90         & 2.69          & 270.30       & 0.81            & 0.57           \\
                           &                       &                      & 2.00         & 2.67          & 281.65       & 0.82            & 0.56           \\
                           &                       &                      & 2.10         & 2.73          & 292.07       & 0.82            & 0.56           \\
            \bottomrule
        \end{tabular}%
    }
    \caption{\textbf{Comparison of models, parameters, epochs, and CFG values.} for model trained on resolution 256×256}
    \label{tab:complete_results_res256}
\end{table*}

%% file: tables/complete_results_res384.tex
\begin{table*}[t]
    \centering
    \resizebox{\linewidth}{!}{%
        \setlength{\tabcolsep}{10pt}
        \renewcommand{\arraystretch}{1.22}
        \footnotesize
        \begin{tabular}{lccccccc}
            \toprule
            \textbf{Model} & \textbf{Params}       & \textbf{Epoch}       & \textbf{CFG} & \textbf{FID↓} & \textbf{IS↑} & \textbf{Prec.↑} & \textbf{Rec.↑} \\
            \midrule\midrule

            \multirow{7}{*}{\textbf{B-384}}
                           & \multirow{7}{*}{120M}
                           & 50                    & 2.00                 & {5.96}         & 190.16        & 0.79         & 0.47                             \\ \cline{3-8}
                           &                       & 100                   & 2.00         & 5.22          & 213.44       & 0.82            & 0.46           \\ \cline{3-8}
                           &                       & 200                  & 2.00         & 4.74          & 230.43       & 0.81            & 0.48           \\ \cline{3-8}
                           &                       & \multirow{4}{*}{300}
                           & 1.75                  & 5.46                 & 196.58       & 0.78          & 0.52                                            \\
                           &                       &                      & 1.90         & 4.62          & 223.31       & 0.81            & 0.50           \\
                           &                       &                      & 2.00         & 4.41          & 237.38       & 0.82            & 0.48           \\
                           &                       &                      & 2.10         & 4.29          & 254.54       & 0.83            & 0.47           \\
            \midrule\midrule

            \multirow{7}{*}{\textbf{L-384}}
                           & \multirow{7}{*}{352M}
                           & 50                    & 2.00                 & {3.43}         & 285.02        & 0.83         & 0.52                             \\ \cline{3-8}
                           &                       & 100                   & 2.00         & 3.10          & 290.11       & 0.82            & 0.53           \\ \cline{3-8}
                           &                       & 200                  & 2.00         & 3.10          & 298.13       & 0.82            & 0.53           \\ \cline{3-8}
                           &                       & \multirow{4}{*}{300}
                           & 1.75                  & 3.00                 & 256.07       & 0.79          & 0.58                                            \\
                           &                       &                      & 1.90         & 2.79          & 283.84       & 0.81            & 0.55           \\
                           &                       &                      & 2.00         & 2.84          & 299.32       & 0.82            & 0.55           \\
                           &                       &                      & 2.10         & 2.93          & 315.02       & 0.83            & 0.54           \\
            \midrule\midrule

            \multirow{6}{*}{\textbf{XL-384}}
                           & \multirow{6}{*}{789M}
                           & 50                    & 2.00                 & {2.98}         & 289.90        & 0.81         & 0.55                             \\ \cline{3-8}
                           &                       & 100                   & 2.00         & 2.77          & 307.30       & 0.82            & 0.56           \\ \cline{3-8}
                           &                       & 200                  & 2.00         & 2.58          & 308.11       & 0.83            & 0.55           \\ \cline{3-8}
                           &                       & \multirow{3}{*}{300}
                           & 1.75                  & 2.81                 & 261.09       & 0.79          & 0.59                                            \\
                           &                       &                      & 1.90         & 2.60          & 285.43       & 0.81            & 0.57           \\
                           &                       &                      & 2.00         & 2.62          & 299.31       & 0.82            & 0.57           \\
                           &                       &                      & 2.10         & 2.68          & 314.75       & 0.82            & 0.56           \\
            \bottomrule
        \end{tabular}%
    }
    \caption{\textbf{Comparison of models, parameters, epochs, and CFG values.} for model trained on resolution 384×384}
    \label{tab:complete_results_res384}
\end{table*}

%% file: tables/complete_results_res_scaling.tex
\begin{table*}[t]
    \centering
    \resizebox{\linewidth}{!}{%
        \setlength{\tabcolsep}{8pt}
        \renewcommand{\arraystretch}{1.2}
        \footnotesize
        \begin{tabular}{lcccccccc}
            \toprule
            \textbf{Model} & \textbf{Epoch} & \textbf{Plen} & \textbf{Threshold} & \textbf{CFG} & \textbf{FID$\downarrow$} & \textbf{IS$\uparrow$} & \textbf{Prec.$\uparrow$} & \textbf{Rec.$\uparrow$} \\
            \midrule\midrule

            \multirow{4}{*}{\textbf{DPAR-B-128}}
                           & \multirow{4}{*}{299}
                           & \multirow{4}{*}{7.60}
                           & \multirow{4}{*}{4.00}
                           & 1.75 & {21.41} & 74.52 & 0.53 & 0.59 \\
                           &  &  &  & 1.90 & 18.48 & 88.29 & 0.56 & 0.57 \\
                           &  &  &  & 2.00 & 16.88 & 97.23 & 0.57 & 0.57 \\
                           &  &  &  & 2.10 & 15.55 & 106.41 & 0.58 & 0.55 \\
            \midrule

            \multirow{4}{*}{\textbf{LlamaGen-B-128}}
                           & \multirow{4}{*}{299}
                           & \multirow{4}{*}{NA}
                           & \multirow{4}{*}{NA}
                           & 1.75 & {24.53} & 63.50 & 0.51 & 0.53 \\
                           &  &  &  & 1.90 & 22.46 & 72.26 & 0.53 & 0.49 \\
                           &  &  &  & 2.00 & 21.56 & 78.81 & 0.54 & 0.49 \\
                           &  &  &  & 2.10 & 20.75 & 84.82 & 0.56 & 0.47 \\
            \midrule

            \multirow{4}{*}{\textbf{DPAR-B-256}}
                           & \multirow{4}{*}{299}
                           & \multirow{4}{*}{7.80}
                           & \multirow{4}{*}{4.00}
                           & 1.75 & {5.02} & 193.99 & 0.78 & 0.54 \\
                           &  &  &  & 1.90 & 4.28 & 219.40 & 0.81 & 0.52 \\
                           &  &  &  & 2.00 & 4.07 & 235.39 & 0.82 & 0.50 \\
                           &  &  &  & 2.10 & 3.98 & 250.62 & 0.83 & 0.49 \\
            \midrule

            \textbf{LlamaGen-B-256} & 299 & NA & NA & 2.00 & {5.46} & 193.61 & 7.50 & 0.84 \\
            \midrule

            \multirow{4}{*}{\textbf{DPAR-B-384}}
                           & \multirow{4}{*}{299}
                           & \multirow{4}{*}{7.90}
                           & \multirow{4}{*}{4.00}
                           & 1.75 & {5.46} & 196.58 & 0.78 & 0.52 \\
                           &  &  &  & 1.90 & 4.62 & 223.31 & 0.81 & 0.50 \\
                           &  &  &  & 2.00 & 4.41 & 237.38 & 0.82 & 0.48 \\
                           &  &  &  & 2.10 & 4.29 & 254.54 & 0.83 & 0.47 \\
            \midrule

            \textbf{LlamaGen-B-384} & 299 & NA & NA & 2.25 & {6.09} & 182.54 & 7.24 & 0.84 \\
            \midrule

            \multirow{4}{*}{\textbf{DPAR-B-512}}
                           & \multirow{4}{*}{299}
                           & \multirow{4}{*}{8.10}
                           & \multirow{4}{*}{4.00}
                           & 1.75 & {8.20} & 165.76 & 0.74 & 0.52 \\
                           &  &  &  & 1.90 & 6.69 & 190.46 & 0.76 & 0.51 \\
                           &  &  &  & 2.00 & 6.05 & 205.51 & 0.78 & 0.49 \\
                           &  &  &  & 2.10 & 5.59 & 222.18 & 0.79 & 0.48 \\
            \midrule

            \multirow{4}{*}{\textbf{LlamaGen-B-512}}
                           & \multirow{4}{*}{299}
                           & \multirow{4}{*}{NA}
                           & \multirow{4}{*}{NA}
                           & 1.75 & {10.61} & 110.63 & 0.76 & 0.48 \\
                           &  &  &  & 1.90 & 9.20 & 124.23 & 0.78 & 0.46 \\
                           &  &  &  & 2.00 & 8.44 & 133.39 & 0.80 & 0.44 \\
                           &  &  &  & 2.10 & 7.97 & 142.48 & 0.81 & 0.42 \\
            \bottomrule
        \end{tabular}%
    }
    \caption{\textbf{Resolution-scaling complete results for DPAR-B and LlamaGen-B.} Results at 128, 256, 384, and 512 resolutions across sampled CFG values.}
    \label{tab:complete_results_res_scaling}
\end{table*}

%% file: visualizations/visualizations.tex
\begin{figure*}
    \includegraphics[width=\linewidth]{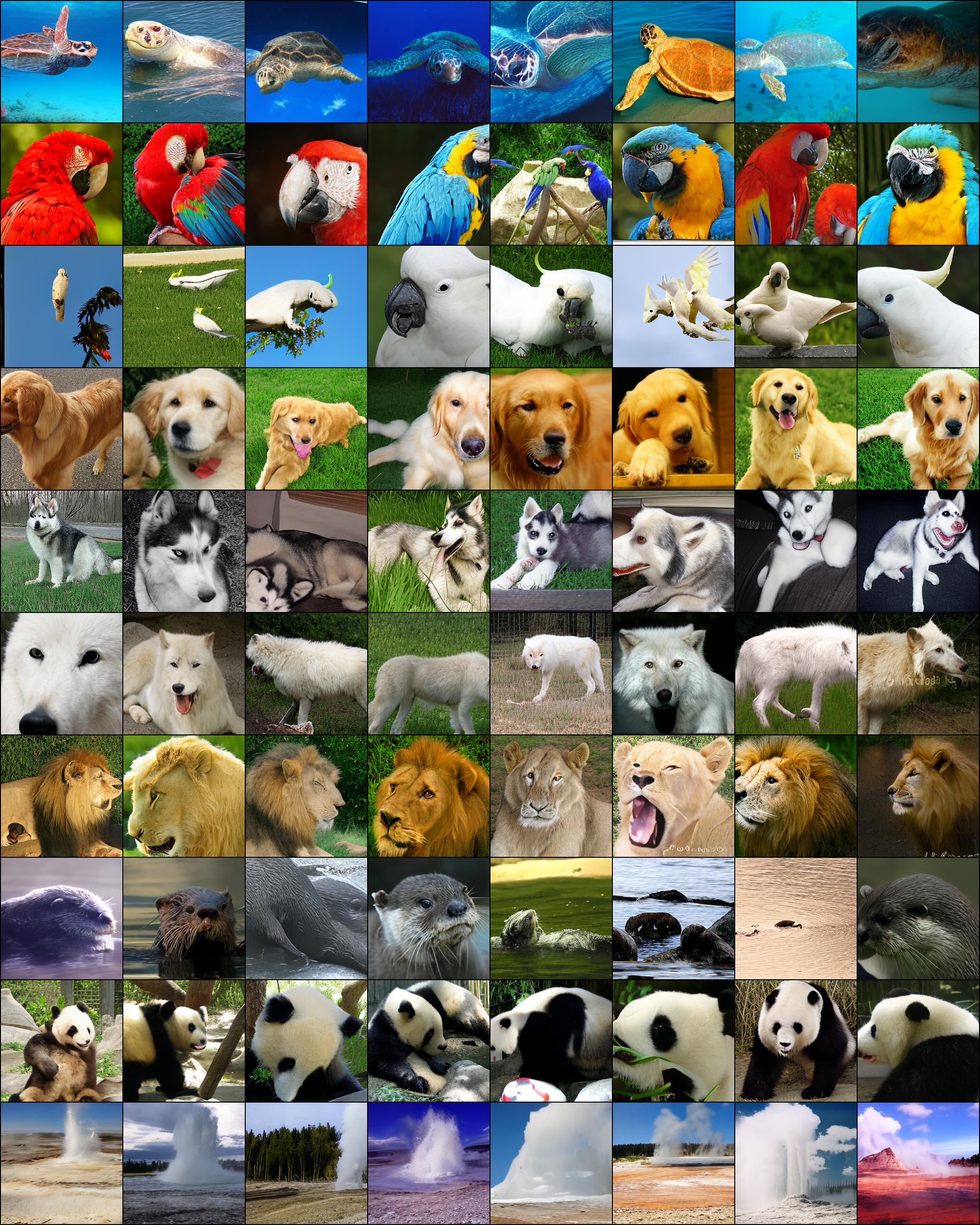}
    \caption{Uncurated generated samples for model \textbf{\algoname{}-XL} trained at 256$\times$256 resolution at \textbf{CFG-scale=1.75}}
\end{figure*}
\begin{figure*}
    \includegraphics[width=\linewidth]{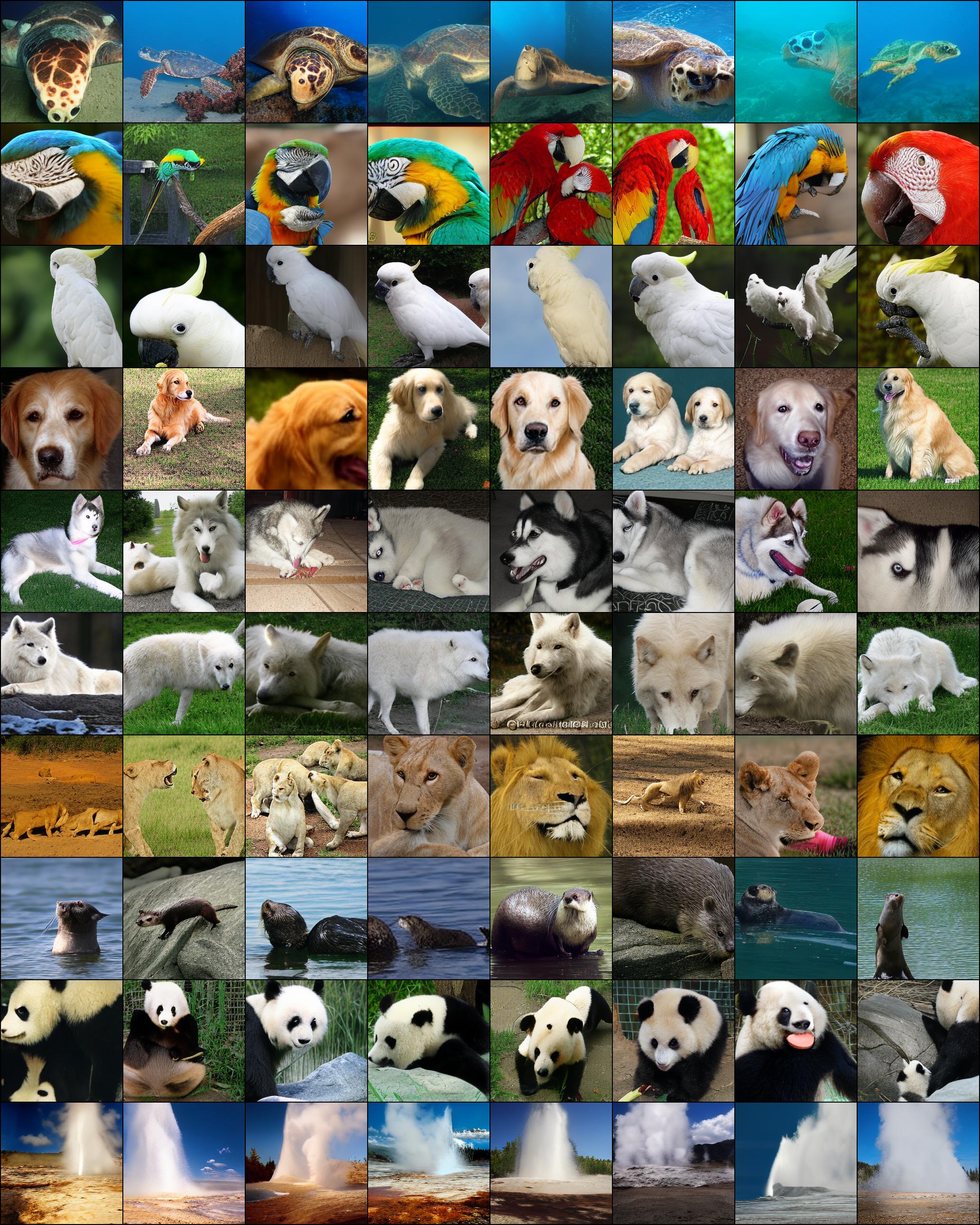}
    \caption{Uncurated generated samples for model \textbf{\algoname{}-XL} trained at 256$\times$256 resolution at \textbf{CFG-scale=1.9}}
\end{figure*}
\begin{figure*}
    \includegraphics[width=\linewidth]{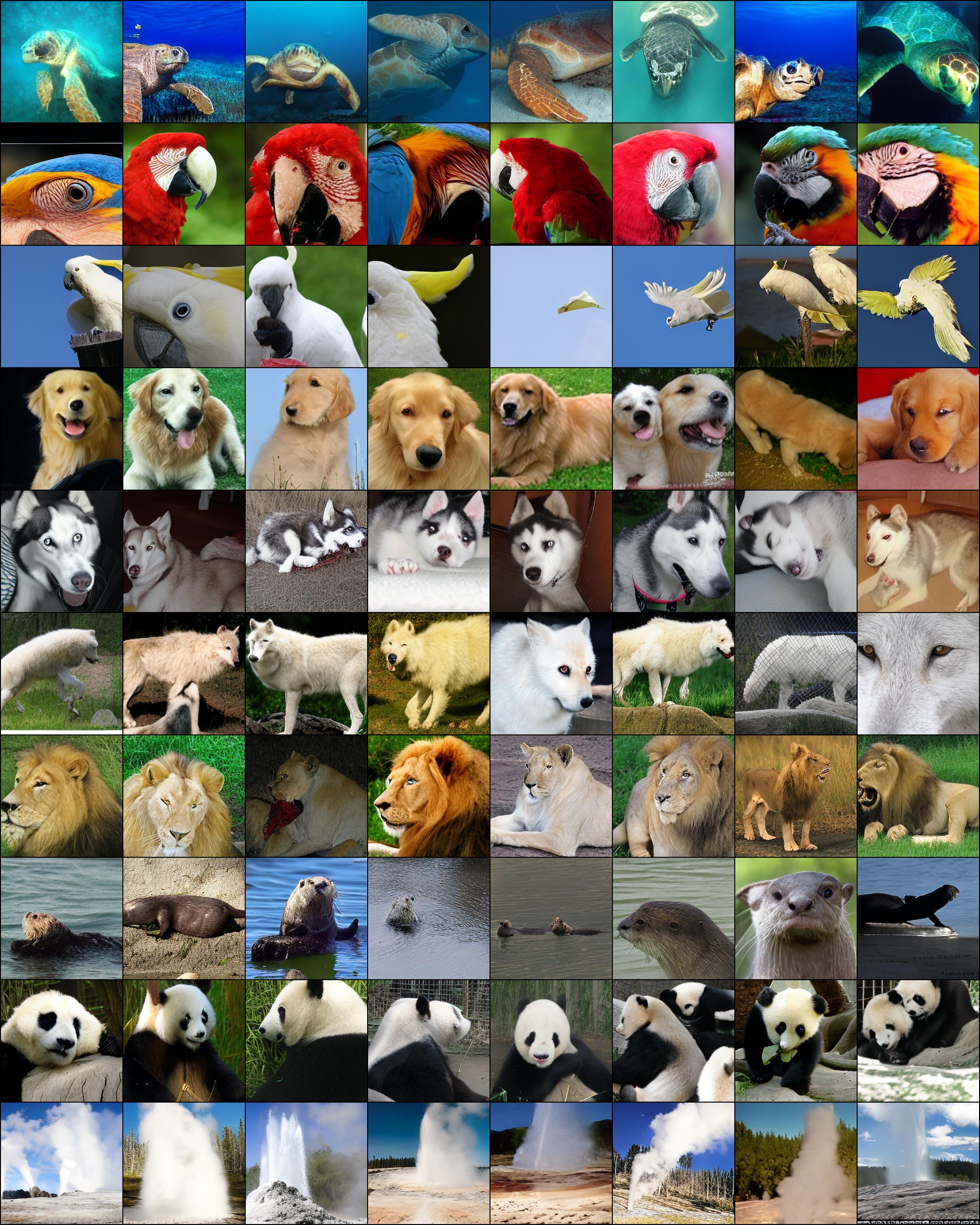}
    \caption{Uncurated generated samples for model \textbf{\algoname{}-XL} trained at 256$\times$256 resolution at \textbf{CFG-scale=2.0}}
\end{figure*}
\begin{figure*}
    \includegraphics[width=\linewidth]{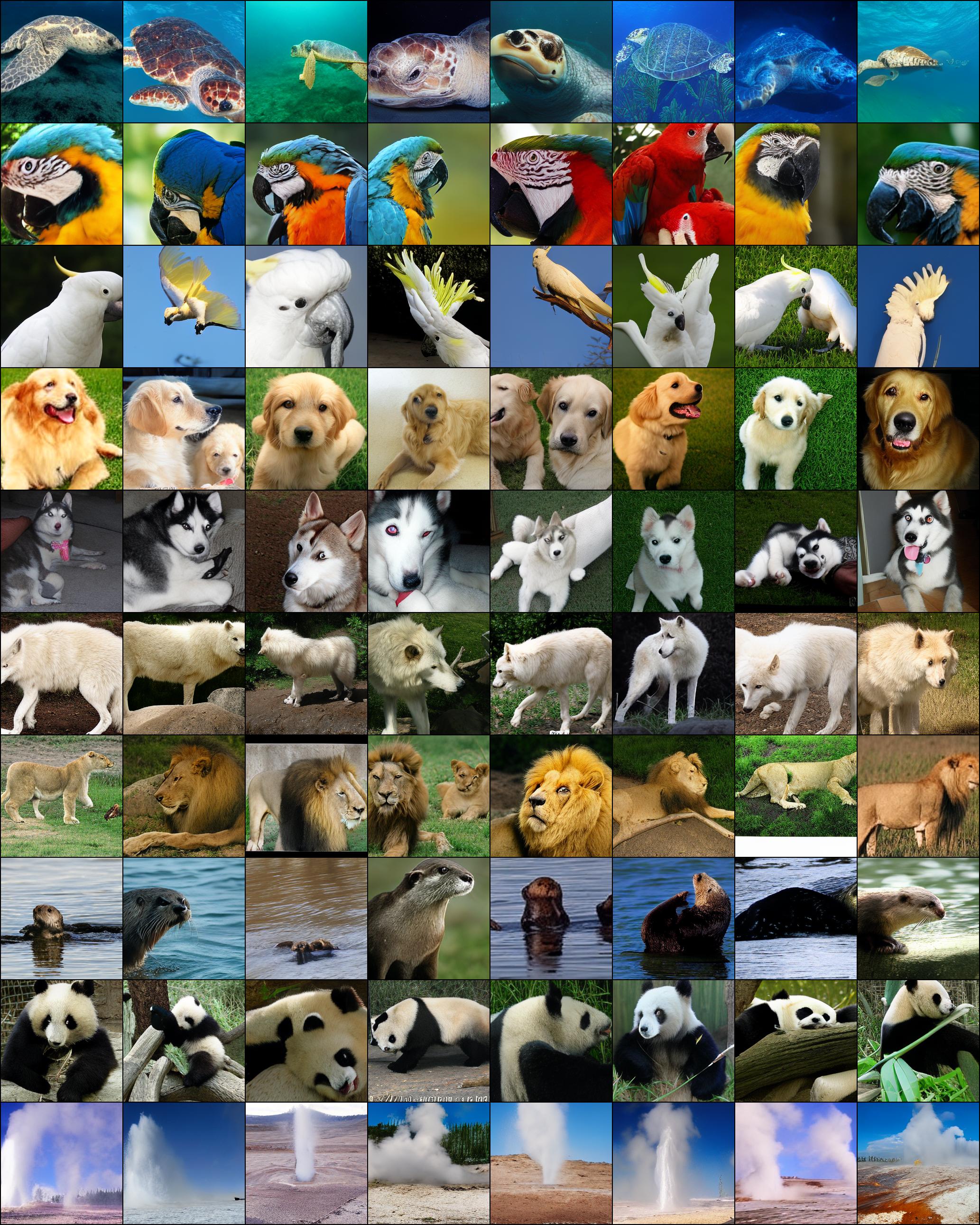}
    \caption{Uncurated generated samples for model \textbf{\algoname{}-XL} trained at 256$\times$256 resolution at \textbf{CFG-scale=2.1}}
\end{figure*}
\begin{figure*}
    \includegraphics[width=\linewidth]{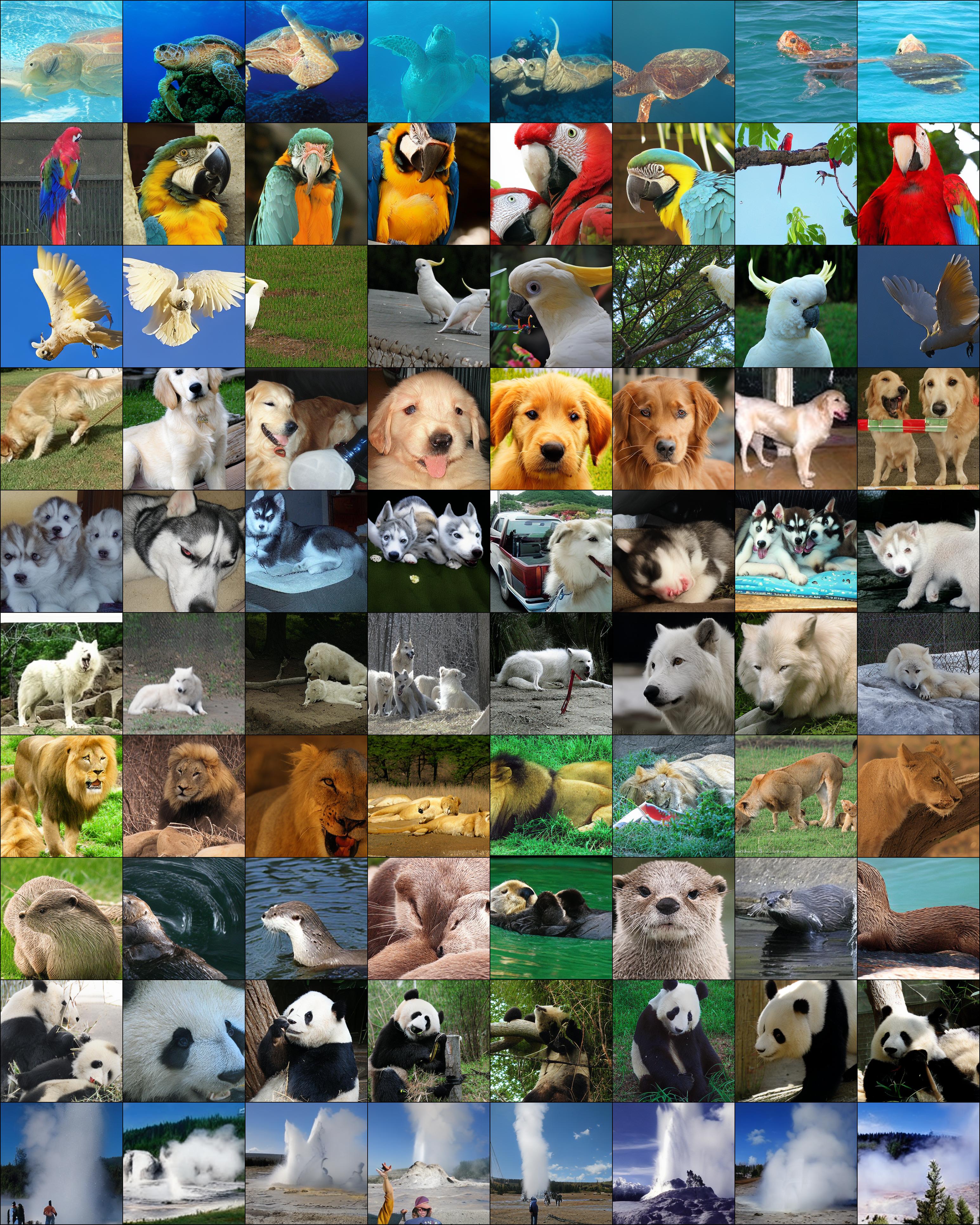}
    \caption{Uncurated generated samples for model \textbf{\algoname{}-XL} trained at 384$\times$384 resolution at \textbf{CFG-scale=1.75}}
\end{figure*}
\begin{figure*}
    \includegraphics[width=\linewidth]{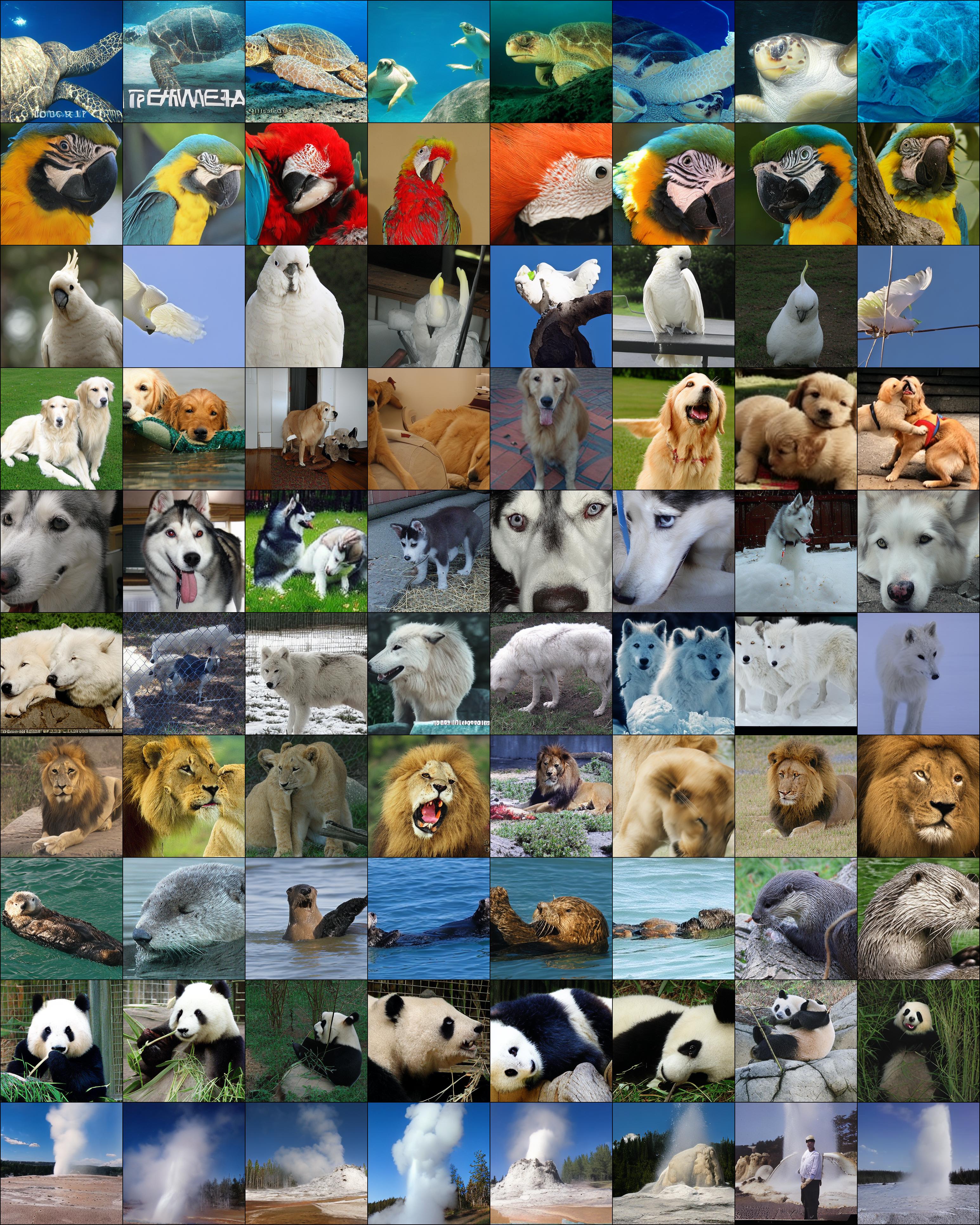}
    \caption{Uncurated generated samples for model \textbf{\algoname{}-XL} trained at 384$\times$384 resolution at \textbf{CFG-scale=1.9}}
\end{figure*}
\begin{figure*}
    \includegraphics[width=\linewidth]{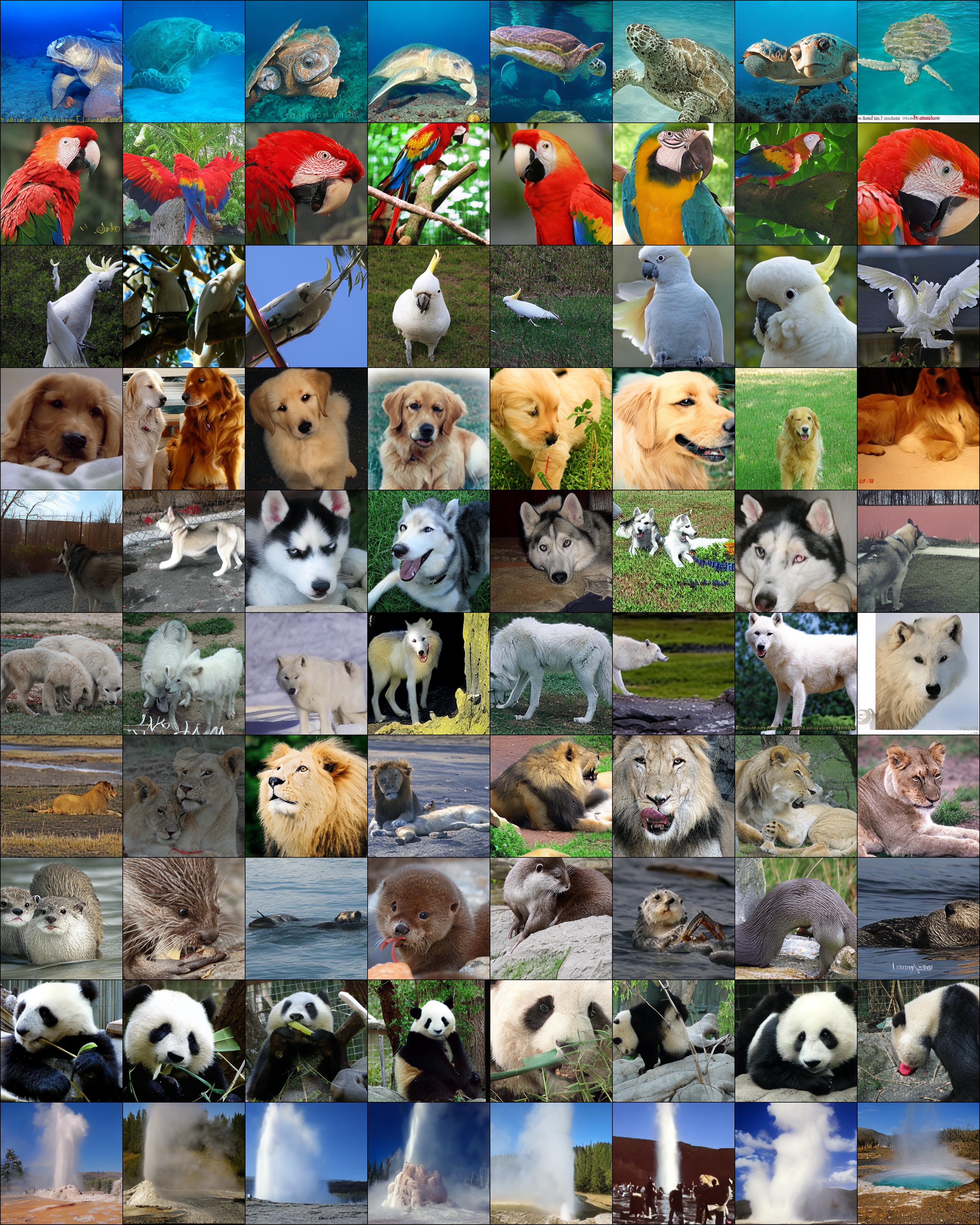}
    \caption{Uncurated generated samples for model \textbf{\algoname{}-XL} trained at 384$\times$384 resolution at \textbf{CFG-scale=2.0}}
\end{figure*}
\begin{figure*}
    \includegraphics[width=\linewidth]{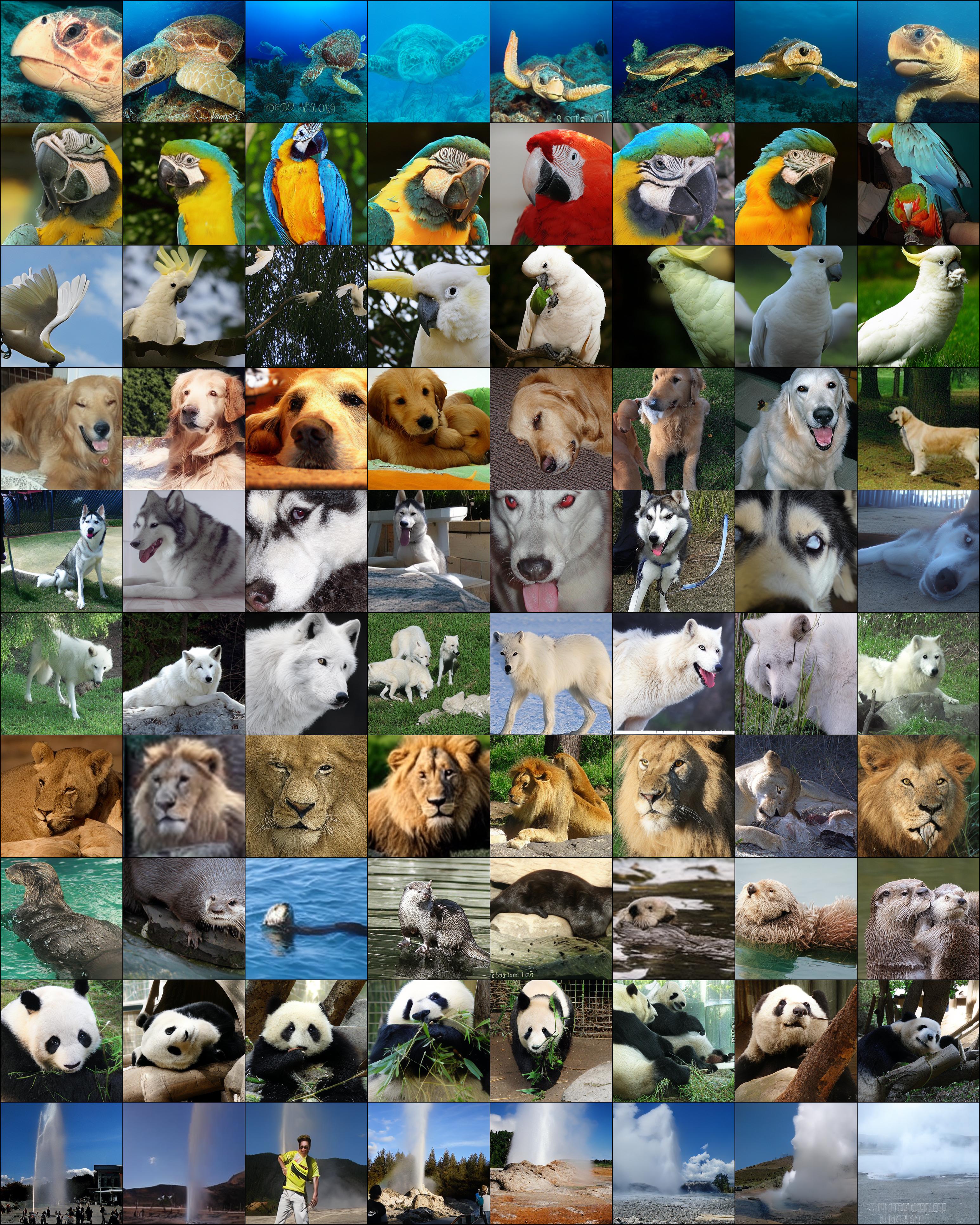}
    \caption{Uncurated generated samples for model \textbf{\algoname{}-XL} trained at 384$\times$384 resolution at \textbf{CFG-scale=2.1}}
\end{figure*}